\documentclass{article}

\PassOptionsToPackage{numbers, compress}{natbib}
\usepackage[preprint]{neurips_2026}

\usepackage[utf8]{inputenc} % allow utf-8 input
\usepackage[T1]{fontenc}    % use 8-bit T1 fonts
\usepackage{hyperref}       % hyperlinks
\usepackage{url}            % simple URL typesetting
\usepackage{booktabs}       % professional-quality tables
\usepackage{amsfonts}       % blackboard math symbols
\usepackage{nicefrac}       % compact symbols for 1/2, etc.
\usepackage{microtype}      % microtypography

\usepackage{pifont}
\usepackage[dvipsnames,table]{xcolor}

\definecolor{lightblue}{RGB}{220,235,250}

\usepackage{graphicx}
\usepackage{subcaption}
\usepackage{amsmath}

\title{How Seemingly Inconsequential Design Choices Dictate Performance of LLMs in Pathology}

\author{%
  Kian R. Weihrauch\textsuperscript{1,2} \quad
  Thomas A. Buckley\textsuperscript{2} \quad
  William Lotter\textsuperscript{2,3} \quad
  Arjun K. Manrai\textsuperscript{2} \\[0.5ex]
  \normalfont
  \textsuperscript{1}Massachusetts Institute of Technology, Cambridge, MA \\
  \textsuperscript{2}Harvard Medical School, Boston, MA \\
  \textsuperscript{3}Dana-Farber Cancer Institute, Boston, MA \\[0.5ex]
  \texttt{Arjun\_Manrai@hms.harvard.edu}
}

\begin{document}

\maketitle

\begin{abstract}
General-purpose large language models (LLMs) are routinely used as baselines when evaluating specialized pathology models on whole-slide images (WSIs). Because WSIs exceed contemporary model context limits, LLM baselines routinely use small, high-magnification patches processed independently via majority voting, without systematic evaluation of seemingly inconsequential design choices such as patch size, patch count, and magnification. Generalist LLMs have consistently underperformed specialized systems, reinforcing the perception that domain-specific training or architectural adaptation is necessary for pathology tasks involving WSIs. Here, we conduct a systematic factorial analysis of four input design factors: inference mode, patch size, magnification, and patch count. We demonstrate that prior studies have overstated the gap between specialized models and general-purpose LLMs by choosing non-optimized input configurations. On the MultiPathQA benchmark, switching to a single balanced configuration (large patches at lower magnification, processed jointly) raises GPT-5 from 15.1\% to 39.5\% on cancer-type classification (TCGA) and from 38.1\% to 62.9\% on organ classification (GTEx). Per-task optimization yields further gains up to 43.9\% (TCGA) and 71.6\% (GTEx). The same configuration generalizes to two other models and to a fully held-out CPTAC cohort, where it improves Gemini~3~Flash by 23.4 percentage points without any task-specific tuning.

\end{abstract}

\section{Introduction}

General-purpose large language models (LLMs), such as GPT-5 \cite{singh2025openaigpt5card}, hold much promise for the future of conversational multi-modal medical AI. One key use case is medical imaging, and specifically pathology. However, whole-slide images (WSIs) introduce a substantial input challenge. Full-resolution WSI reach gigapixel scale, far exceeding the context window of current models. This has motivated a growing body of work on specialized architectures: pathology-specific vision encoders that compress
WSIs into compact representations~\cite{chen2025slidechatlargevisionlanguageassistant, alawode2026mllm, liang2025wsillavamultimodallargelanguage}, and agentic frameworks that navigate slides through iterative zoom-in strategies~\cite{chen2025pathagentinterpretableanalysiswholeslide, sun2025cpathagentagentbasedfoundationmodel,
buckley2025navigatinggigapixelpathologyimages}. These approaches are consistently benchmarked against general-purpose LLMs, which serve as a key baseline for quantifying the value of domain-specific design.

\begin{figure}[t]
    \centering
    \includegraphics[width=0.85\linewidth]{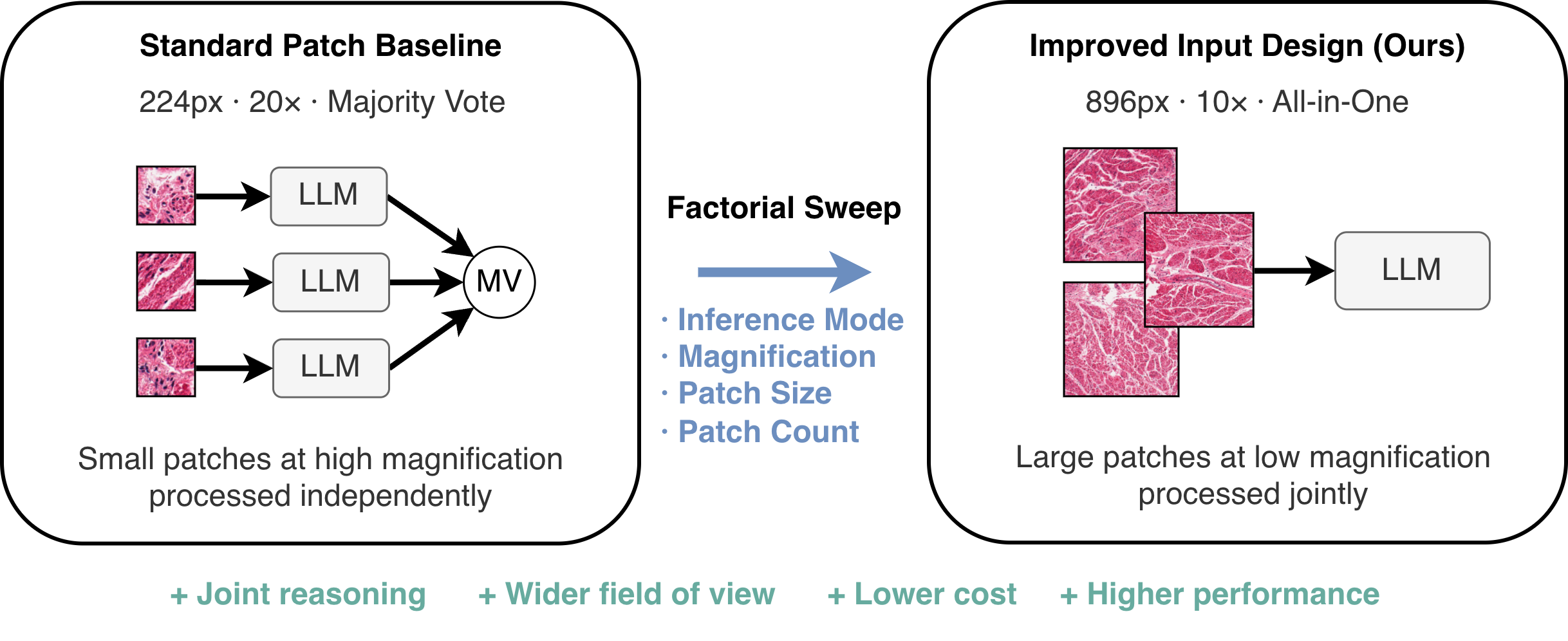}
    \caption{General-purpose LLMs are widely used as baselines for pathology-specific models, yet patching and inference mode have received little attention. Prior work evaluated these models using small, high-magnification patches distributed across independent instances with majority voting. We conduct a factorial experiment over four input design factors: inference mode, patch size, magnification, and patch count. We find that performance is highly sensitive to input configuration. Presenting large, low-magnification patches jointly to a single model instance substantially improves accuracy while reducing inference cost.}
    \label{fig:overview}
\end{figure}

Recent work has largely converged on a small set of evaluation protocols for evaluating general-purpose LLMs as WSI-level baselines: either presenting the model with a low-resolution thumbnail of the full slide, or randomly sampling small, high-magnification patches (e.g. $224 \times 224$ pixel patches at $20\times$ magnification), processing each patch independently, and aggregating predictions through majority voting~\cite{chen2025slidechatlargevisionlanguageassistant,
buckley2025navigatinggigapixelpathologyimages,
chen2025pathagentinterpretableanalysiswholeslide,
liang2025wsillavamultimodallargelanguage,
li2025coevolvingagenticaimedical}. Both protocols lead to poor general-purpose LLM performance. The resulting gap between domain-adapted models is frequently interpreted as evidence that specialized architectures are superior. However, if the baseline itself is suboptimal, the magnitude of this gap and the conclusions drawn from it may be overstated.

Patch-based evaluation generally outperforms thumbnail-based evaluation~\cite{buckley2025navigatinggigapixelpathologyimages,
chen2025pathagentinterpretableanalysiswholeslide,
chen2025slidechatlargevisionlanguageassistant}. However, the patching protocol for general-purpose LLMs itself has not been systematically studied. Several of the design choices for the current patching method are questionable on first principles. First, processing patches independently prevents the model from directly comparing distant regions or jointly reasoning over global tissue structure, potentially limiting the cross-region evidence synthesis required for WSI analysis. Second, patches of size $224 \times 224$ pixels provide only a narrow field of view and may exclude diagnostically important contextual information. Third, although $20\times$ magnification is standard for human pathologists, it is unclear whether small, high-magnification pathology patches align well with the visual distributions seen during pretraining of general-purpose vision encoders. Such inputs likely bear little resemblance to the pathology content these models likely encountered during training, such as textbook figures and journal illustrations that depict tissue architecture at a broader scale.

In this work, we investigate the extent to which input configuration shapes the pathology performance of general-purpose LLMs. We conduct a full-factorial experiment over four key design factors (patch size, magnification, patch count, and inference mode), evaluating 72 configurations of GPT-5 on the MultiPathQA benchmark~\cite{buckley2025navigatinggigapixelpathologyimages}, which spans organ classification, cancer-type classification, cancer grading, and visual question answering (see Fig. \ref{fig:overview}).
Our results show that these input choices have a substantial effect on downstream accuracy, often rivaling or exceeding gains typically attributed to architectural changes. The main findings are:

\begin{enumerate}
    \item \textbf{Input configuration is a dominant source of performance variation.} Changing how patches are presented to the model shifts GPT-5 accuracy from 15.1\% to 43.9\% on cancer-type classification and from 38.1\% to 71.6\% on organ classification. These gains exceed those typically attributed to domain-specific architectural adaptation in prior work.

    \item \textbf{Optimal configurations are task-dependent, and no single setting dominates.} Classification tasks requiring tissue-level context benefit most from larger patches at lower magnification, while tasks requiring cellular detail (e.g., cancer grading) favor higher magnification. A joint inference mode consistently helps, but the remaining factors interact with task demands. Ideally, input configuration should be adjusted based on the task.

    \item \textbf{A balanced configuration consistently improves performance across models and data.} While per-task optimization yields the largest gains, a single balanced setting (896\,px, $10\times$, 20 patches, All-in-One) consistently outperforms the standard protocol across three independently trained models (GPT-5, Qwen~3.5~Plus, Gemini~3~Flash) and generalizes to a held-out benchmark (CPTAC), with gains of up to +23.4 percentage points (pp).
\end{enumerate}

Together, these results suggest that prior general-purpose baselines have been impacted by suboptimal input design. Future evaluations should optimize and report on input configuration before concluding on the superiority of domain-specific approaches.

% \begin{figure}
%     \centering
%     \begin{subfigure}[t]{0.8\linewidth}
%         \centering
%         \includegraphics[width=\linewidth]{figures/PatchProjectFig1.drawio (2).png}
%         \caption{}
%         \label{fig:fig1}
%     \end{subfigure}
%     \centering
%     \begin{subfigure}[t]{0.9\linewidth}
%         \centering
%         \includegraphics[width=\linewidth]{figures/prior_vs_opt_bar_plot.png}
%         \caption{}
%         \label{fig:fig1_bar}
%     \end{subfigure}
% \end{figure}

\section{Related work}

\textbf{Whole-slide image analysis.}
Computational pathology has been dominated by multiple instance learning (MIL), which treats a WSI as a bag of patches and learns to aggregate them into a slide-level prediction. Key milestones include attention-based MIL~\cite{ilse2018attention}, CLAM~\cite{lu2021data}, TransMIL~\cite{shao2021transmil}, and HIPT~\cite{chen2022scaling}. These works demonstrate how the aggregation strategy (e.g., max pooling or attention weighting) critically determines downstream
performance. 
%In practice, MIL pipelines typically operate on a randomly sampled or filtered subset of tissue patches rather than the full slide~\cite{}, making patch selection and coverage an implicit design choice.
Recent work on pathology-specific foundation models has further advanced patch-level feature extraction. Large-scale vision encoders such as UNI~\cite{chen2024towards}, CHIEF~\cite{wang2024pathology}, GigaPath \cite{xu2024whole}, and
Virchow~\cite{vorontsov2024foundation} provide pretrained embeddings that most modern MIL pipelines are built upon. These approaches have been extended to pathology vision-language models (VLMs), including
CONCH~\cite{lu2024visual}, TITAN~\cite{ding2025multimodal}, and PRISM~\cite{shaikovski2024prismmultimodalgenerativefoundation}, which align visual and textual representations for pathology.
Our work includes the examination of a related aggregation question in the LLM setting:
whether a general-purpose model should reason over patches jointly (similar to attention-based MIL) or independently (similar to majority pooling), and how the resolution and scale of those patches affect performance.

\textbf{Vision-language models in pathology.}
Early adaptations of LLMs in pathology focused on analyzing localized regions of interest (ROIs). Systems such as PathChat~\cite{lu2024multimodal}, Quilt-LLaVA~\cite{seyfioglu2024quilt}, PathGen-LLaVA~\cite{sun2025pathgen}, and more recently Patho-R1-7B~\cite{zhang2026patho} achieve strong performance on localized pathology tasks through domain-specific training. However, these models cannot natively process gigapixel whole-slide images (WSIs), which are central to comprehensive clinical decision-making.
Recent specialized systems address WSI-scale pathology through novel input pipelines. One line of work compresses slide content into learned visual representations before passing them to an LLM, as in SlideChat~\cite{chen2025slidechatlargevisionlanguageassistant}, WSI-LLaVA~\cite{liang2025wsillavamultimodallargelanguage}, CPath-Omni \cite{sun2025cpath}, and MLLM-HWSI~\cite{alawode2026mllm}. Another line seeks to mimic the pathologist workflow via iterative navigation, sequentially querying regions at varying magnifications. Representative examples include CPathAgent~\cite{sun2025cpathagentagentbasedfoundationmodel}, SlideSeek~\cite{chen2025evidence},  PathFinder~\cite{ghezloo2025pathfinder}, GIANT~\cite{buckley2025navigatinggigapixelpathologyimages}, Pathology-CoT~\cite{wang2025pathologycotlearningvisualchainofthought}, and PathAgent~\cite{chen2025pathagentinterpretableanalysiswholeslide}. Other multimodal frameworks, such as MedGemma 1.5~\cite{sellergren2026medgemma}, handle WSIs by sampling large grids of low-magnification patches (e.g., up to 126 patches at 5$\times$ magnification) to preserve global context.

\textbf{General-purpose LLMs as WSI baselines.}
To evaluate pathology-specific architectures, recent studies routinely benchmark against general-purpose LLMs (e.g., GPT- and Gemini-class models) using a narrow set of zero-shot protocols. A common baseline evaluates models on low-resolution thumbnails of the full slide~\cite{chen2025slidechatlargevisionlanguageassistant,liang2025wsillavamultimodallargelanguage,chen2025pathagentinterpretableanalysiswholeslide, buckley2025navigatinggigapixelpathologyimages, li2025coevolvingagenticaimedical}. A generally stronger baseline uses randomly sampled patch sets, predicting on small crops independently and aggregating predictions~\cite{chen2025slidechatlargevisionlanguageassistant,buckley2025navigatinggigapixelpathologyimages,chen2025pathagentinterpretableanalysiswholeslide}. This aggregation is likely motivated both by prior work showing model improvements with self-consistency \cite{wang2022self}, and by model limits on the number of allowed input images. However, different from self-consistency, where each model would see the same images, the current approach distributes images, meaning the model never jointly observes multiple regions.
Other works evaluate general-purpose models using hand-selected ROIs~\cite{chen2025evidence}, textual slide descriptions~\cite{ghezloo2025pathfinder, sun2025cpath}, or agentic wrappers that allow autonomous patch selection~\cite{sun2025cpathagentagentbasedfoundationmodel,wang2025pathologycotlearningvisualchainofthought}.

\section{Methods}

\subsection{Experimental setup}

\textbf{Factorial design.}
To systematically analyze the impact of visual input configuration on model performance, we conduct a factorial study over four key factors: inference mode, patch size, magnification, and number of patches.
\textit{Patch size} determines the pixel dimensions of each extracted region from the WSI at a given magnification. \textit{Magnification} controls the resolution level at which patches are sampled from the WSI. Therefore, patch size and magnification govern the field of view captured per patch. \textit{Patch count} specifies the maximum number of random regions extracted per slide. \textit{Inference mode} defines how extracted patches are presented to the model. We consider two modes:
(1)~\textit{Majority Vote}, in which each patch is sent to an independent model instance, and predictions are aggregated via majority
voting~\cite{chen2025slidechatlargevisionlanguageassistant,
buckley2025navigatinggigapixelpathologyimages,
chen2025pathagentinterpretableanalysiswholeslide}, and (2)~\textit{All-in-One}, in which all patches are provided jointly as input to a single model instance. We sweep over magnifications 5$\times$, 10$\times$, 20$\times$, patch sizes 224, 512, 896, 1024\,px, patch counts 10, 20, 30, and the two outlined inference modes. This yields a total of 72 configurations.

% All factor levels are summarized in Table~\ref{tab:factorial}.

% \begin{table}[htbp]
% \caption{Factorial design. The full combination of all factor levels yields 72
% unique configurations.}
% \label{tab:factorial}
% \centering
% \begin{tabular}{ll}
% \toprule
% \textbf{Factor} & \textbf{Levels} \\
% \midrule
% Inference Mode & Majority Vote, All-in-One \\
% Magnification & 5$\times$, 10$\times$, 20$\times$ \\
% Patch Size & 224, 512, 896, 1024\,px \\
% Patch Count & 10, 20, 30 \\
% \bottomrule
% \end{tabular}
% \end{table}

\textbf{Dataset.}
Experiments are conducted on the MultiPathQA
benchmark~\cite{buckley2025navigatinggigapixelpathologyimages}, which spans five datasets with a total of 934 questions. MultiPathQA covers both classification and visual question answering (VQA) tasks. The classification tasks include organ classification (GTEx \cite{gtex}), organ--cancer classification (TCGA \cite{weinstein2013cancer}), and cancer grading (PANDA \cite{pandachallenge}). The VQA section includes questions from SlideBench \cite{chen2025slidechatlargevisionlanguageassistant} and the pathologist authored ExpertVQA \cite{buckley2025navigatinggigapixelpathologyimages}, which require free-text answers to questions about WSI content. We select this benchmark for its diversity of task types and its documented difficulty: the paper reports limited performance for both general-purpose and specialized models~\cite{buckley2025navigatinggigapixelpathologyimages}. We use the same metrics and prompting outlined in the MultiPathQA paper.

\textbf{Patch extraction and inference.}
Patches are created using Trident~\cite{zhang2025standardizing}. For each WSI and each magnification--patch size combination, we randomly sample 30 patches. Subsets of 10 and 20 patches are drawn from this set of 30, ensuring that smaller patch counts are strict subsets of larger ones. This nested design isolates the effect of patch count without introducing additional sampling variability.
We treat the specified patch count as a maximum. For a small number of slides, particularly in the PANDA dataset, low magnification combined with a large patch size results in fewer patches than requested (details in the Appendix \ref{app:patching}). In these cases, all available patches are used, effectively covering the entire slide. We retain these samples in the analysis, as full-slide coverage is itself a possible benefit stemming from a combination of patch size, magnification, and patch count.

All factorial experiments use the GPT-5 API via Azure. OpenAI API constraints impose a maximum of 50 images and 50\,MB per request. The maximum resolution of GPT-5 before down-sampling occurs is $1024 \times 1024$ pixels. These constraints motivate the upper bounds of our factor levels: patch count is capped at 30 (50 patches consistently resulted in larger than 50MB requests) and patch size at 1024\,px. As we input medical data, the Azure API blocked a small number of requests due to input content violations (details in Appendix \ref{app:refusal}). We removed these samples from the downstream analysis.
Patch sizes range from $224 \times 224$\,px, the setting most commonly used in prior work, to $1024 \times 1024$\,px, with intermediate values of 512 and 896\,px. Magnification levels are set to 5$\times$, 10$\times$, and 20$\times$; higher levels are excluded because PANDA slides are predominantly scanned at 20$\times$.
For Majority Vote inference, patch predictions are generated independently for all 30 patches per slide. Results for smaller patch counts (10 and 20) are computed from subsets of these predictions, avoiding redundant API calls. Cost and token usage for all experiments are reported in the Appendix \ref{app:cost_token}.

\subsection{Evaluation strategy}

To reduce the amount of API calls, we adopt a two-phase evaluation strategy designed to first identify the most relevant factors on a subset of slides, and then to investigate the most task-dependent factors and their interactions over the entire dataset.

\textbf{Phase 1: Full factorial over MultiPathQA subset.}
We first run the complete 72 configuration factorial on a stratified subset of 100 WSIs (20 randomly sampled slides/questions per dataset). Because we expect task-dependent dynamics, organ classification, cancer grading, and VQA pose fundamentally different visual challenges, we analyze each benchmark separately, averaging results over the 20 slides per factor combination.
To quantify the relative contribution of each factor, we perform a two-way interaction ANOVA per benchmark, estimating effect sizes with omega squared ($\omega^2$). We note that the reuse of slides, along with the derivation of Majority Vote subsets from shared runs, violates the independence assumption of ANOVA. Accordingly, we use it to identify which factors explain the most variance relative to one another, rather than to test for statistical significance.

\textbf{Phase 2: Full-scale evaluation.}
Guided by Phase~1, which identified inference mode as the dominant factor, we conduct two experiments on the complete 934-question MultiPathQA benchmark. First, to validate the preliminary finding at scale, we compare All-in-One and Majority Vote using the standard configuration from prior literature (224$\times$224\,px patches, 20$\times$ magnification, 30 patches), isolating the effect of inference mode while holding all other factors constant. Second, having confirmed the superiority of All-in-One, we run a full factorial over the remaining three factors (patch size, magnification, and patch count) using only the All-in-One inference mode, yielding 36 configurations evaluated across all slides. As the Majority Vote configuration uses many more reasoning tokens than All-in-One inference, excluding Majority Vote drastically reduces the cost of the experiments.

\textbf{Phase 3: Cross-model and held-out evaluation.}
To test whether the scaling trends identified in Phase~2 generalize beyond GPT-5 and the MultiPathQA benchmarks, we conduct a separate evaluation along two axes: new models and held-out data. We evaluate two additional general-purpose models (Qwen 3.5 Plus \cite{qwen3.5} and Gemini 3 Flash \cite{pichai2025new}) via the OpenRouter API. Gemini 3 Flash was selected because OpenRouter ranked it as the most-used model for health-related queries at the time of evaluation\footnote{\href{https://openrouter.ai/rankings?category=health\#categories}{OpenRouter health rankings}}. As a held-out benchmark, we construct a 12-way cancer-type classification task from 200 slides sampled from the Clinical Proteomic Tumor Analysis Consortium (CPTAC)~\cite{edwards2015cptac}; 15 slides that could not be converted to TIFF format or tiled by Trident \cite{zhang2025standardizing} were excluded, yielding 185 evaluation samples. This dataset was not used in any prior analysis or configuration selection.
For each model we compare two All-in-One configurations: a
\textit{naive} setting mirroring prior work (224\,px, 20$\times$, 30 patches) and an \textit{optimized} setting (896\,px, 10$\times$, 20 patches), chosen to balance performance across tasks by selecting factor levels near observed plateaus rather than optimizing for any individual benchmark. Majority Vote is excluded because Phase~2 establishes its consistent inferiority and because it incurs approximately 10$\times$ higher per-sample cost, making multi-model evaluation at scale prohibitively expensive.

\section{Results}

\subsection{Exploratory factorial experiment to prioritize factors}

\begin{figure}[h!]
    \centering
    \includegraphics[width=0.85\linewidth]{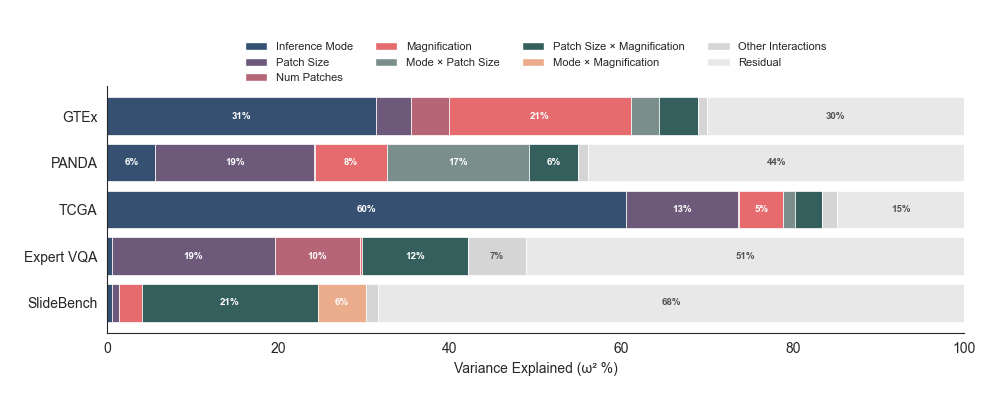}
    \caption{Variance decomposition of GPT-5 results on 100 WSI subset using ANOVA with two-factor interactions. Percentages are shown for factors explaining more than 5\% of the variance.}
    \label{fig:variance_decomp_subset}
\end{figure}

\begin{figure}[h!]
    \centering
    \includegraphics[width=0.8\linewidth]{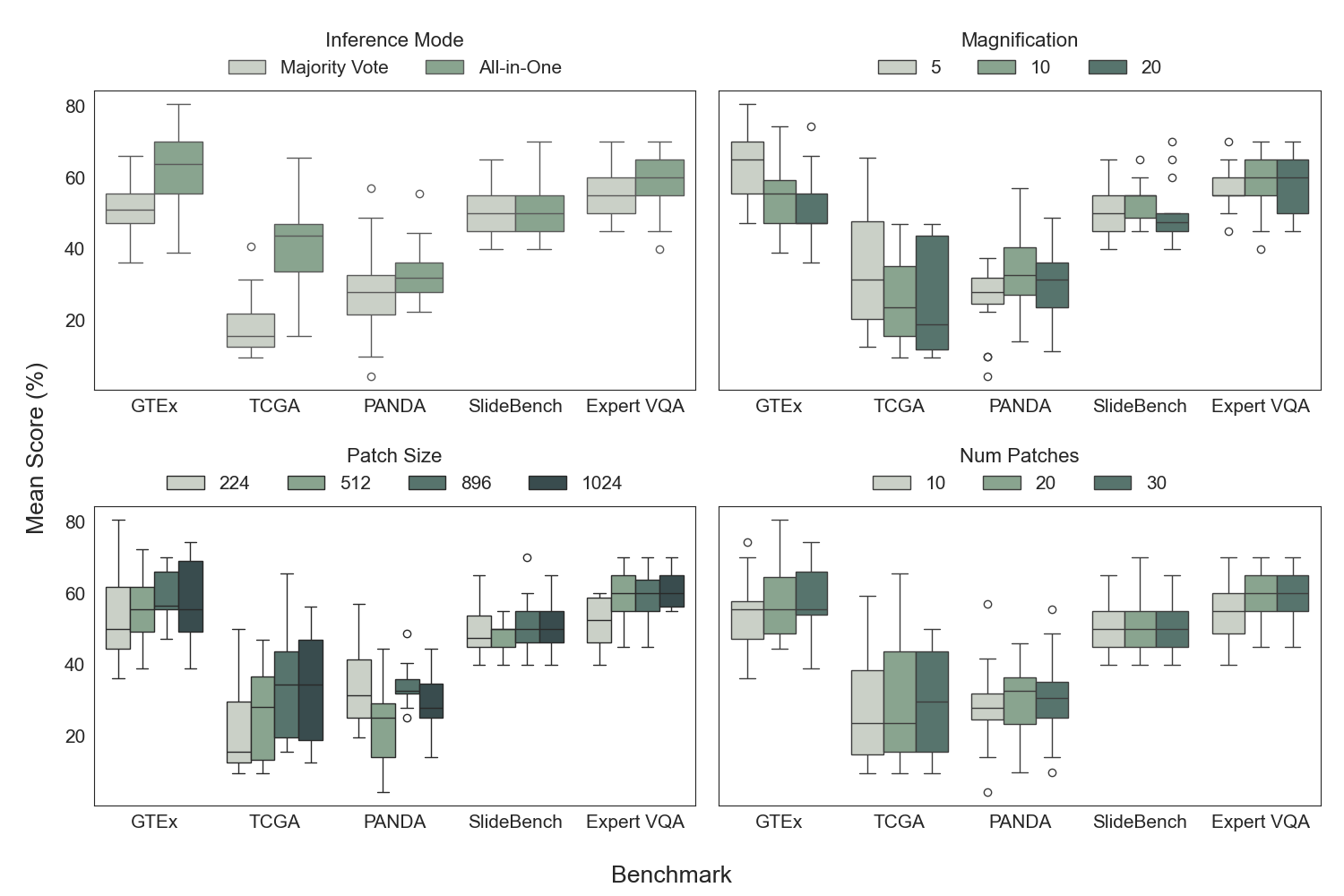}
    \caption{Performance distribution on the 100-WSI subset, grouped by factor level. Each box aggregates over all configurations sharing that level.}
    \label{fig:boxplots_subset_new}
\end{figure}

We first conducted an exploratory factorial sweep over all four factors on a randomly sampled subset of 100 WSIs (20 per dataset). With only 20 slides per benchmark, this phase is intended to prioritize which factors warrant full-scale evaluation, not to draw definitive conclusions.
Variance decomposition via two-way interaction ANOVA (Fig.~\ref{fig:variance_decomp_subset}) suggests task-dependent dominant factors. However, we find that All-in-One inference may benefit performance across tasks (Fig.~\ref{fig:boxplots_subset_new}) and could account for a substantial share of variance, particularly on classification tasks (Fig.~\ref{fig:variance_decomp_subset}).
Increasing patch size and decreasing magnification seem to lead to additional gains on GTEx, TCGA, and VQA tasks, while patch count contributes minimally (Fig.~\ref{fig:boxplots_subset_new}).
Interaction effects seem to be generally small; the most consistent being \textit{Patch Size~$\times$~Magnification}. PANDA is an exception to this, as \textit{Mode~$\times$~Patch Size} seems to account for large parts of variance. This is driven by a performance drop under Majority Vote at larger patch sizes while All-in-One remains stable (Appendix Fig.~\ref{fig:bottom5_mag}).
Based on these patterns, we hypothesize that All-in-One inference yields broad performance improvements across tasks. Because we examine factor effects across five benchmarks and multiple interaction terms, individual benchmark-specific patterns at this sample size are vulnerable to noise mining. We therefore carry forward only the effect that is largest in magnitude and consistent across benchmarks (inference mode), and treat all other Phase~1 patterns as exploratory.

\subsection{Validation: All-in-One vs.\ Majority Vote at scale}

\begin{figure}[h!]
    \centering
    \begin{subfigure}[t]{0.45\linewidth}
        \centering
        \includegraphics[width=\linewidth]{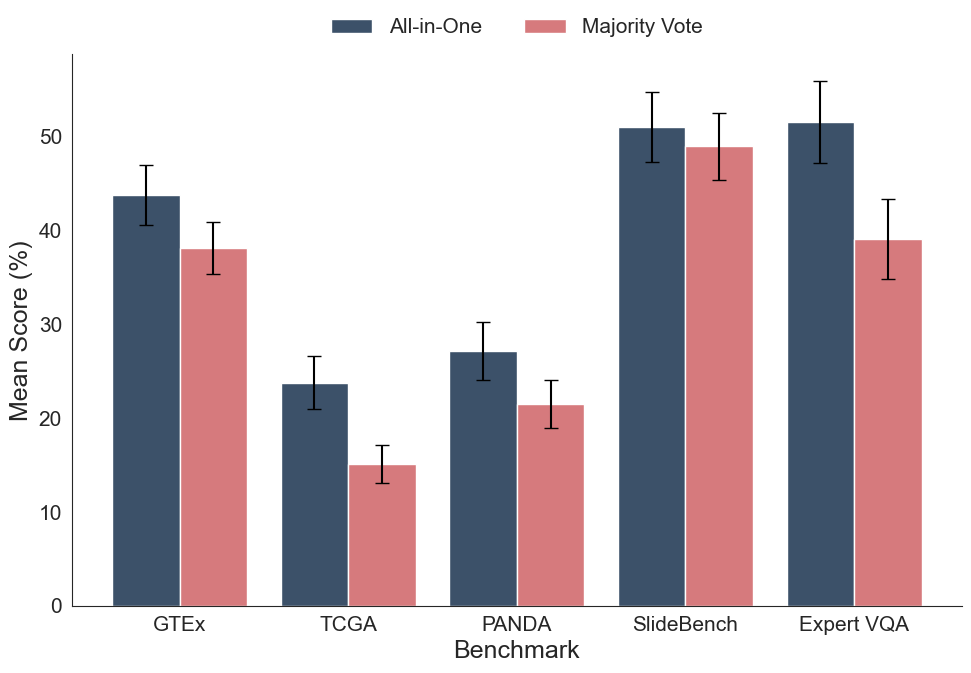}
        \caption{Inference Mode Performance Comparison}
        \label{fig:inference_comp}
    \end{subfigure}
    \hfill
    \begin{subfigure}[t]{0.45\linewidth}
        \centering
        \includegraphics[width=\linewidth]{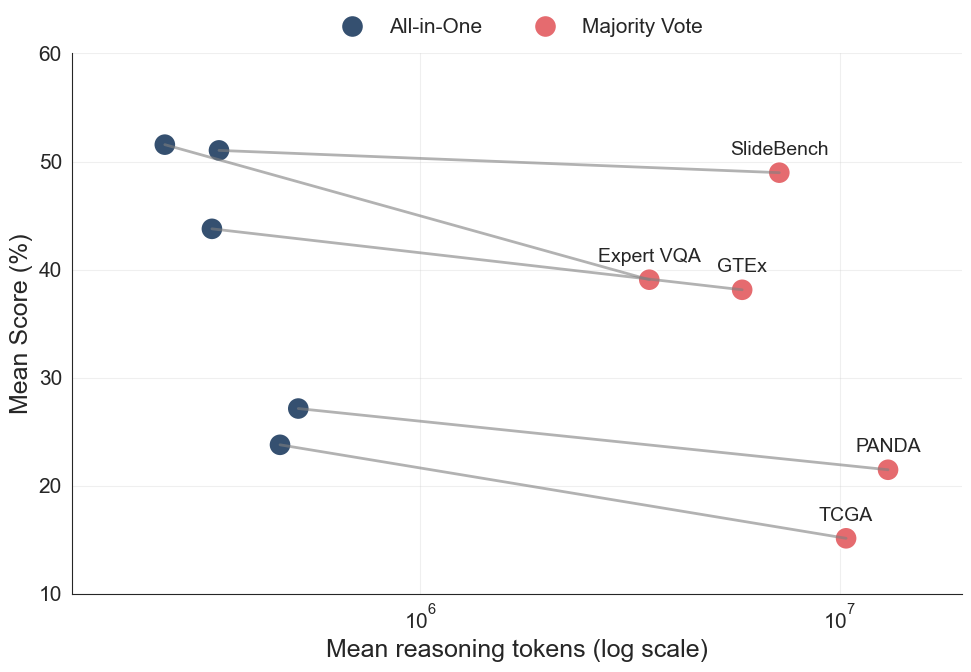}
        \caption{Inference Mode Cost Comparison}
        \label{fig:cost_performance}
    \end{subfigure}
    \caption{All-in-One vs.\ Majority Vote on the full 934-question MultiPathQA dataset. Both inference modes used 30 patches at 20$\times$ magnification with 224$\times$224\,px patch size. Scores are balanced accuracy for classification tasks and accuracy for VQA tasks. Error bars show standard deviation from a 1000-sample bootstrap.
    (a) All-in-One outperforms Majority Vote across all tasks.
    (b) All-in-One incurs substantially lower compute cost by reasoning jointly over patches rather than independently.}
    \label{fig:full_inference_comp}
\end{figure}

To confirm the effect of inference mode in a controlled setting, we compare All-in-One and Majority Vote on the full MultiPathQA benchmark using the patch settings from prior work (224$\times$224,px, 20$\times$, 30p). As shown in Figure~\ref{fig:inference_comp}, the All-in-One mode outperforms Majority Vote across all five benchmarks. While we initially expected only modest differences on the VQA tasks, ExpertVQA exhibits the largest improvement (+12.50 pp), alongside substantial gains on the classification benchmarks GTEx (+5.64 pp), TCGA (+8.65 pp), and PANDA (+5.67 pp). On SlideBench, All-in-One achieves a modest improvement (+2.06 pp). Because a single model call reasons jointly over all patches, it avoids the redundant context processing inherent to Majority Vote, where each patch requires independent inference calls and reasoning (Fig.~\ref{fig:cost_performance}).

\subsection{Full-scale analysis: scaling behavior and interactions}

We next evaluated all remaining factors (patch size, magnification, and patch count) in the All-in-One setting across the complete MultiPathQA benchmark.

\begin{figure}[h!]
    \centering
    \includegraphics[width=0.9\linewidth]{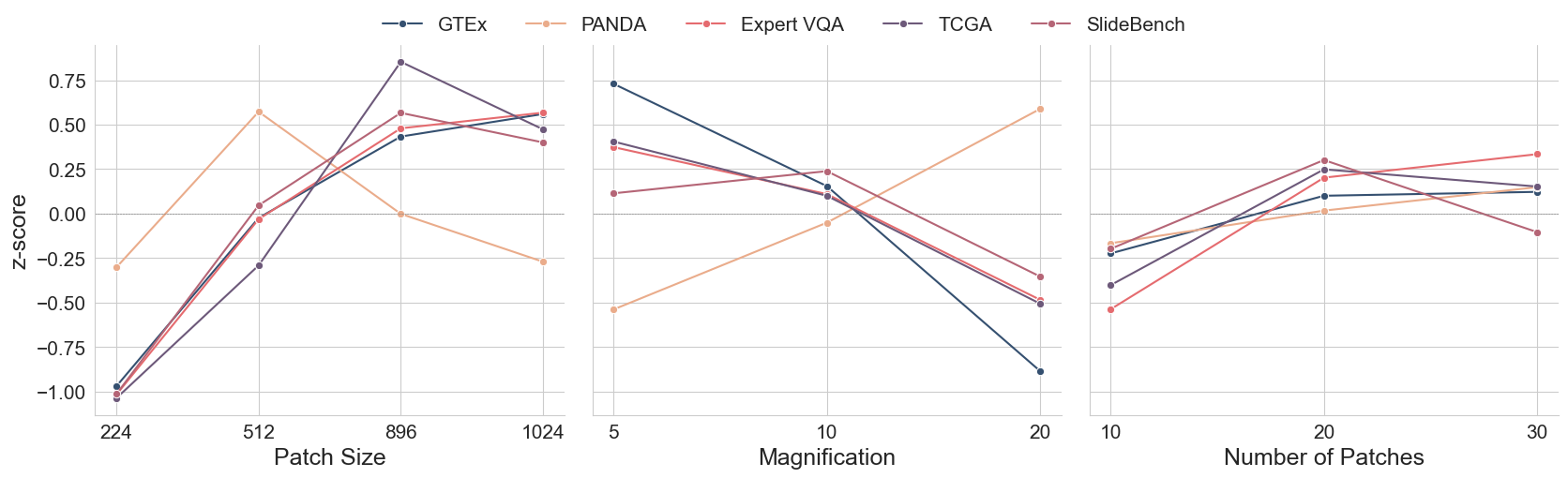}
    \caption{Scaling curves under All-in-One inference across the full 934-question MultiPathQA dataset. Performance (z-score normalized per benchmark) is shown as a function of patch size (left), magnification (center), and number of patches (right).}
    \label{fig:scaling_curves}
\end{figure}

\textbf{Scaling curves.}
Figure~\ref{fig:scaling_curves} shows z-score--normalized performance as a function of each factor. Two consistent trends emerge across most benchmarks: (1) performance improves with increasing patch size, and (2) performance improves with decreasing magnification (i.e., larger field of view per patch). Increasing patch count yields diminishing returns, with performance generally plateauing at 20 patches. The PANDA dataset is a notable exception: it achieves optimal performance at an intermediate patch size (512\,px) and benefits from higher magnification (10--20$\times$), likely reflecting the finer cellular detail required for Gleason grading compared to organ- or cancer-type classification.

\begin{figure}[h!]
    \centering
    \includegraphics[width=0.95\linewidth]{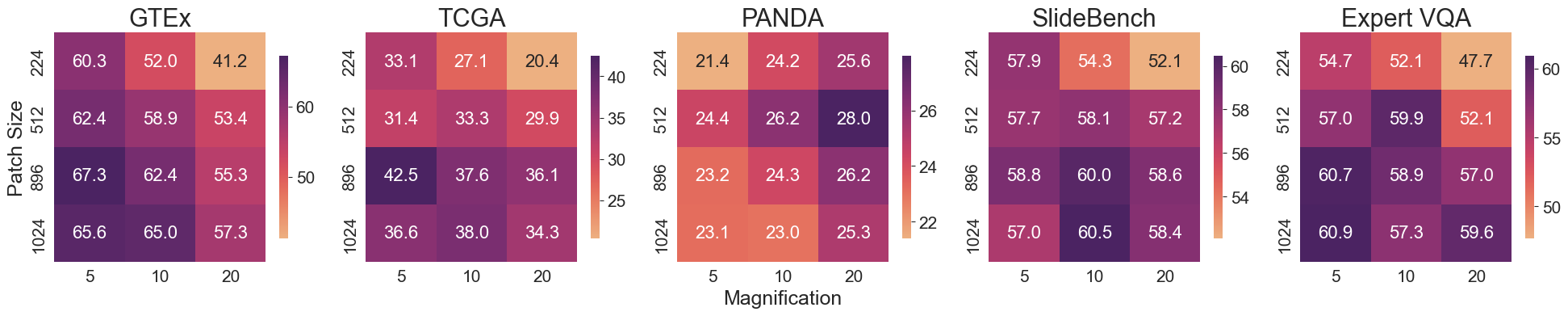}
    \caption{Heatmaps showing accuracy as a function of patch size and magnification under All-in-One inference, averaged over patch count. Each cell represents the mean over the full 934-question MultiPathQA dataset for a given benchmark.}
    \label{fig:heatmap_full}
\end{figure}

\textbf{Patch size--magnification interaction.}
As patch count yields small effects, we focus on examining the effects of magnification and patch size. The heatmaps in Figure~\ref{fig:heatmap_full} reveal the joint effect of patch size and magnification on each benchmark. For GTEx and TCGA, the highest accuracies cluster in the low-magnification, large-patch-size regime (e.g., 1024\,px at 5$\times$), consistent with these tasks relying primarily on tissue-level architectural features. PANDA shows a flatter landscape with a slight preference for high magnification (20$\times$) and somewhat lower patch sizes (512px). SlideBench benefits from larger patch sizes at 10$\times$, and ExpertVQA seems to favor low magnification with a large patch size.

\begin{figure}[h!]
    \centering
    \includegraphics[width=0.85\linewidth]{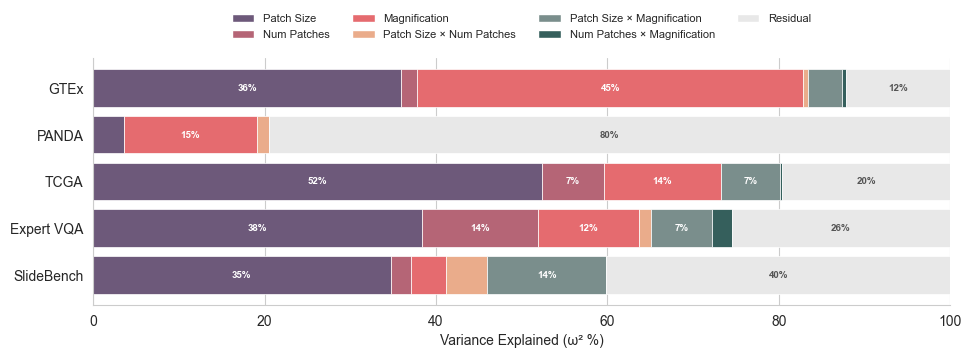}
    \caption{Variance decomposition over the full 934-question MultiPathQA dataset (All-in-One inference only) using ANOVA with two-factor interactions. Percentages are shown for factors explaining more than 5\% of the variance.}
    \label{fig:full_decomp}
\end{figure}

\textbf{Variance decomposition.}
A two-way ANOVA over the full dataset (Fig.~\ref{fig:full_decomp}) confirms that \textit{Patch Size} and \textit{Magnification} are the dominant main effects, with largely additive contributions: their interaction exceeds 5\% only for TCGA (7\%), ExpertVQA (7\%), and SlideBench (14\%). Residuals are smallest for GTEx (12\%) and TCGA (20\%), indicating input configuration accounts for most performance variation on classification tasks, while PANDA (80\%) and SlideBench (40\%) are dominated by task-intrinsic variability. This reflects that the model struggles with Gleason grading regardless of input configuration, and that the heterogeneity of VQA questions can introduce variability that no single input setting can resolve.

\textbf{Mixed Magnification Ablation.} We tested whether combining patches from multiple magnifications within a single input could capture complementary spatial information; however, mixed magnification inputs did not consistently improve over fixed-magnification settings (Appendix~\ref{app:mixed_mag}).

\textbf{Field of view Ablation.}
Patch size, magnification, and patch count jointly determine the total field of view (FoV) captured per slide, raising the question of whether their effects are determined by FoV alone. We fit benchmark-specific logarithmic curves relating total FoV to performance and measure how performance deviates from these curves based on each factor (Appendix~\ref{app:fov}). FoV explains some of the scaling behavior, but benchmark-specific residuals persist: GTEx continues to favor lower magnification, PANDA higher magnification, and TCGA and GTEx remain sensitive to patch size beyond what FoV alone predicts. This indicates that patch size, magnification, and patch count are not merely proxies for tissue coverage, but that they shape the visual representation provided to the model in partially independent ways.

\subsection{Comparison with prior work}

\begin{table}[t]
\centering
\caption{Performance on the full 934-question MultiPathQA benchmark. Results for prior approaches are taken from the original MultiPathQA paper~\cite{buckley2025navigatinggigapixelpathologyimages}. \textit{Agent} refers to the GIANT framework. Our results report the best-performing GPT-5 configuration identified for each benchmark vs. our results running the prior work protocol.}
\label{tab:simplified_wsi}
\small
\setlength{\tabcolsep}{6pt}
\setlength{\tabcolsep}{5pt}
\begin{tabular}{llccccc}
\toprule
Model & Mode & TCGA & GTEx & PANDA & SB VQA & ExpertVQA \\
\midrule

\multicolumn{7}{l}{\textbf{General-purpose models (prior work)}} \\
GPT-5 & Patch
& $15.5 \pm 2.2$
& $42.3 \pm 2.4$
& $21.1 \pm 2.4$
& $51.5 \pm 3.6$
& $41.4 \pm 4.4$ \\
GPT-5 & Thumbnail
& $9.2 \pm 1.9$
& $36.2 \pm 3.3$
& $12.4 \pm 2.3$
& $54.0 \pm 3.6$
& $50.0 \pm 4.4$ \\
GPT-5 & Agent
& $32.3 \pm 3.5$
& $54.1 \pm 3.2$
& $23.4 \pm 2.3$
& $58.0 \pm 3.5$
& $57.0 \pm 4.5$ \\
\midrule

\multicolumn{7}{l}{\textbf{Pathology-trained WSI Chat Models (prior work)}} \\
SlideChat & WSI
& $3.3 \pm 1.2$
& $4.7 \pm 0.3$
& $17.0 \pm 0.3$
& $\mathbf{70.5 \pm 3.2}$
& $37.5 \pm 4.3$ \\
MedGemma 1.5 & WSI
& $6.2 \pm 1.5$
& $24.3 \pm 1.8$
& $19.6 \pm 2.5$
& $31.7 \pm 3.3$
& $35.9 \pm 4.2$ \\
PathAgent & WSI
& $9.3 \pm 1.7$
& $38.9 \pm 3.2$
& $19.2 \pm 2.5$
& $49.5 \pm 3.6$
& $44.5 \pm 4.5$ \\
\midrule

\multicolumn{7}{l}{\textbf{Our evaluation}} \\
GPT-5 & Prior Protocol & $15.1 \pm 2.0$ & $38.1 \pm 2.7$ & $21.5 \pm 2.5$ & $49.0 \pm 3.6$ & $39.1 \pm 4.3$ \\
GPT-5 & Best Config. & $\mathbf{43.9 \pm 3.9}$ & $\mathbf{71.6 \pm 2.9}$ & $\mathbf{31.0 \pm 2.9}$ & $61.9 \pm 3.4$ & $\mathbf{63.3 \pm 4.3}$ \\
\midrule
\multicolumn{2}{l}{$\Delta$ (Best vs.\ Prior Protocol)} & $+28.8$ & $+33.5$ & $+9.5$ & $+12.9$ & $+24.2$ \\
\bottomrule
\end{tabular}
\end{table}

Table~\ref{tab:simplified_wsi} places our best-performing GPT-5 input configurations in the context of results reported in the original MultiPathQA benchmark \cite{buckley2025navigatinggigapixelpathologyimages}.
For each benchmark, we report the strongest configuration identified in our factorial evaluation. We also report our results using the protocol from prior work for direct comparison. We find that under optimized visual input configurations, GPT-5 substantially improves over the standard Majority Vote patch baseline used in prior work~\cite{chen2025slidechatlargevisionlanguageassistant,
buckley2025navigatinggigapixelpathologyimages,
chen2025pathagentinterpretableanalysiswholeslide}. On TCGA and GTEx, performance rises from 15.1\% to 43.9\% and from 38.1\% to 71.6\%, respectively. On PANDA, gains from input optimization are more modest, consistent with our variance decomposition showing that configuration choices explain less variance for this benchmark.
We report the best-performing configuration settings for each benchmark in Appendix~\ref{app:impr_over_prior}.
Most pathology-specialized models evaluated on MultiPathQA perform poorly on classification tasks, making direct comparison less informative. The more meaningful reference points are SlideChat, which dominates SlideBench VQA  (70.5\%), and GIANT (Agent), which achieves strong results on ExpertVQA (57.0\%). The standard GPT-5 patch baseline previously trailed GIANT, but our optimized configuration closes and exceeds this gap, purely through input design changes.

\subsection{Generalization to new models and held-out data}

To evaluate whether the scaling trends identified above generalize beyond the specific model and benchmarks used to derive them, we test along two axes: (1)~new model providers and (2)~a held-out dataset not used during configuration selection. Importantly, both configurations already use All-in-One inference, which Phase~2 established as uniformly superior to Majority Vote. The gains reported here arise purely from patch configuration choices (size, magnification, count).
As shown in Table~\ref{tab:model_suite}, the patch-optimized configuration consistently improves accuracy across TCGA, GTEx, ExpertVQA, and SlideBench for all three models. On TCGA and GTEx, gains range from +15.0 to +31.9 percentage points. Performance on PANDA remains largely unchanged, indicating that it continues to be a challenging benchmark. Critically, the same pattern holds on the held-out CPTAC benchmark, which played no role in configuration selection: accuracy improves by +3.8 (GPT-5), +7.2 (Qwen~3.5~Plus), and +23.4 (Gemini~3~Flash) percentage points.
The consistency of these effects across three independently trained models from different providers, as well as on a fully held-out classification task, suggests that the scaling trends in Figure~\ref{fig:scaling_curves} reflect an intrinsic property of how these models operate, rather than model- or dataset-specific effects.

\begin{table}[h!]
\centering
\caption{Generalization of the patch-optimized configuration to new models and a held-out benchmark (CPTAC, shaded) on the entire 934-question MultiPathQA dataset. The naive setting mirrors prior work (All-in-One, 224\,px, 20$\times$, 30 patches); the optimized setting uses 896\,px, 10$\times$, 20 patches.}
\label{tab:model_suite}
\small
\setlength{\tabcolsep}{5pt}
\resizebox{\linewidth}{!}{
\begin{tabular}{llccccc>{\columncolor{lightblue}}c}
\toprule
Model & Config & TCGA & GTEx & PANDA & SlideBench & ExpertVQA & \cellcolor{lightblue} CPTAC (val) \\
\midrule
GPT-5 & Naive & $23.8 \pm 2.8$ & $43.8 \pm 3.2$ & $27.1 \pm 3.1$ & $49.0 \pm 3.6$ & $39.1 \pm 4.3$ & \cellcolor{lightblue} $32.7 \pm 2.7$ \\
GPT-5 & Optimized & $39.5 \pm 3.4$ & $62.9 \pm 2.9$ & $28.9 \pm 3.0$ & $61.9 \pm 3.4$ & $60.2 \pm 4.3$ & \cellcolor{lightblue} $36.5 \pm 3.1$ \\
\midrule
Qwen 3.5 Plus & Naive & $32.0 \pm 3.1$ & $50.8 \pm 3.4$ & $19.9 \pm 2.2$ & $57.9 \pm 3.6$ & $54.7 \pm 4.3$ & \cellcolor{lightblue} $32.5 \pm 3.1$ \\
Qwen 3.5 Plus & Optimized & $54.3 \pm 3.3$ & $66.8 \pm 3.0$ & $18.9 \pm 1.4$ & $58.4 \pm 3.5$ & $60.9 \pm 4.3$ & \cellcolor{lightblue} $39.7 \pm 3.3$ \\
\midrule
Gemini 3 Flash & Naive & $31.1 \pm 3.4$ & $60.5 \pm 3.6$ & $26.0 \pm 3.1$ & $56.3 \pm 3.6$ & $56.2 \pm 4.2$ & \cellcolor{lightblue} $32.5 \pm 3.1$ \\
Gemini 3 Flash & Optimized & $63.0 \pm 3.2$ & $75.5 \pm 3.0$ & $27.6 \pm 3.4$ & $60.9 \pm 3.4$ & $64.8 \pm 4.1$ & \cellcolor{lightblue} $55.9 \pm 3.4$ \\
\bottomrule
\end{tabular}
}
\end{table}

\section{Conclusion}

We demonstrate that patching and inference mode are dominant but largely overlooked determinants of general-purpose LLM performance on WSI tasks. Simply changing how patches are presented to the model, without modifying the model itself, yields gains of up to +33.5 percentage points over prior evaluation protocols. While optimal patching settings are task-dependent, a single balanced configuration (All-in-One inference, 896\,px, $10\times$, 20 patches) performs robustly across models and generalizes to held-out data. Our findings suggest that prior work may have overstated the performance gap between general-purpose LLM baselines and domain-specific systems due to suboptimal input design choices. We encourage future work to explicitly tune and report input design choices to ensure fair, transparent, and reproducible comparisons.

\textbf{Broader Impacts.} Our findings suggest that general-purpose LLMs may be stronger baselines than previously assumed, motivating improved evaluation practices. While their accessibility may enable stronger baselines and broader use in medical imaging, it also raises risks of application to sensitive clinical data without sufficient validation.

\textbf{Limitations.}
Our claims are based on several assumptions. Our factorial analysis assumes that the four factors studied capture the primary axes of input design variation. Other factors, such as patch overlap or image preprocessing, are not explored. Results are derived from proprietary API-based models evaluated at a specific point in time. Model updates may alter absolute performance, although we expect relative patterns to be more stable.
Several additional limitations stem from API inference costs. While MultiPathQA offers task diversity, optimizing configurations on the same benchmark used for evaluation risks overstating generalization. We mitigate this risk by including a held-out CPTAC evaluation, though broader validation across additional benchmarks remains needed. Full factorial analysis was conducted only for GPT-5. For Qwen 3.5 Plus and Gemini 3 Flash, we compared only two configurations, leaving open the possibility that different models have different optimal settings. Finally, the prohibitive cost of Majority Vote inference (approximately 10× per sample) prevented a full-scale factorial analysis that includes both inference modes.

\bibliographystyle{plainnat}
\bibliography{references}

%%%%%%%%%%%%%%%%%%%%%%%%%%%%%%%%%%%%%%%%%%%%%%%%%%%%%%%%%%%%
\newpage
\appendix
\begin{center}
    \Large\textbf{Appendix}
\end{center}

\section{Implementation and evaluation details}
\subsection{Patching details}
\label{app:patching}
Whole-slide image (WSI) patches were extracted using Trident~\cite{zhang2025standardizing}. Trident delivers patch coordinates. For each slide and each magnification--patch size configuration, we randomly sampled up to 30 patch coordinates using a deterministic shuffle with a fixed seed, ensuring that smaller patch-count settings were strict subsets of larger ones. This allowed for the patch count to be varied without introducing additional sampling variability. 
Patches were extracted directly from WSIs with OpenSlide, using the Trident coordinates. We evaluated patch sizes of 224, 512, 896, and 1024\,px at magnifications of 5$\times$, 10$\times$, and 20$\times$.
Small amounts of slide settings resulted in patch counts under 10, 20, and 30 patches across most benchmarks, letting models see the entire image with less than 30 patches. PANDA exhibited a substantially higher proportion of low-patch slide-settings, particularly for the fewer-than-30-patch threshold (34.2\%; Table~\ref{tab:low_patch_stats}).

\begin{table}[h!]
\centering
\caption{
Distribution of slide-settings with low patch counts across benchmarks.
Each slide-setting corresponds to a unique combination of image,
magnification, and patch size. Percentages are computed relative to the
total number of valid slide-settings within each benchmark.
}
\label{tab:low_patch_stats}
\begin{tabular}{lrrrrrrr}
\toprule
Benchmark & Total & $<10$ & \%$<10$ & $<20$ & \%$<20$ & $<30$ & \%$<30$ \\
\midrule
ExpertVQA & 1536 & 0   & 0.00 & 11  & 0.72 & 18  & 1.17 \\
GTEx      & 2292 & 13  & 0.57 & 49  & 2.14 & 88  & 3.84 \\
PANDA     & 2364 & 151 & 6.39 & 503 & 21.28 & 809 & 34.22 \\
SlideBench& 2328 & 5   & 0.21 & 44  & 1.89 & 77  & 3.31 \\
TCGA      & 2652 & 8   & 0.30 & 39  & 1.47 & 76  & 2.87 \\
\bottomrule
\end{tabular}
\end{table}

\subsection{API details}
Inference for the full-factorial experiments was performed using GPT-5 via both the Azure OpenAI Batch API and the standard Azure OpenAI API. For the exploratory subset factorial experiment, the standard API was used for All-in-One inference, whereas the asynchronous Batch API was used for Majority Vote inference due to the substantially larger token requirements associated with independent patch-level predictions. The second factorial experiment on the full MultiPathQA dataset was conducted exclusively using the Batch API.

All requests were issued using the Responses API with high image detail and high reasoning effort, and detailed reasoning summaries enabled. Inputs consisted of a single user message containing the task prompt and one or more image patches encoded as base64 image URLs. No system prompt was used. For All-in-One inference, all sampled patches were provided within a single request, while for Majority Vote inference, each patch was processed independently and predictions were aggregated via majority voting. OpenRouter experiments were conducted using the same inference configuration, including identical prompts, image inputs, reasoning settings, and aggregation procedures, ensuring comparability across providers.

API calls were executed with automatic retry logic (up to 3 attempts) using exponential backoff for transient errors (e.g., rate limits, timeouts). Parallelization was used for Majority Vote inference (for real-time API), with up to 10 concurrent requests per slide. Token usage, latency, and error statistics were recorded for all requests.
Cost, token usage, and error rates are discussed in Section~\ref{app:cost_token_error}. We report the average wall-clock time per WSI for All-in-One inference on the MultiPathQA subset (see Table \ref{tab:aio_duration}). All other factorial experiments were conducted using the asynchronous Azure Batch API, for which wall-clock time depends on queueing and scheduling and is therefore not directly comparable across runs. The Azure Batch API guarantees a turnaround time of less than 24 hours.

\begin{table}[h!]
\centering
\caption{Average All-in-One inference duration per WSI on the MultiPathQA subset. Duration is measured as end-to-end wall-clock time per request.}
\label{tab:aio_duration}
\begin{tabular}{lrr}
\toprule
Benchmark & Samples & Avg. duration / WSI (s) \\
\midrule
GTEx & 720 & 23.59 \\
PANDA & 720 & 35.49 \\
TCGA & 720 & 29.47 \\
TCGA ExpertVQA & 720 & 25.62 \\
TCGA SlideBench & 720 & 28.11 \\
\bottomrule
\end{tabular}
\end{table}

\subsection{Evaluation details}
Model outputs were evaluated using a custom evaluation pipeline that parsed predictions from raw GPT-5 responses and matched them against the MultiPathQA ground-truth annotations. For all benchmarks, the model had structured multiple-choice outputs; the evaluation script extracted answers from JSON-formatted responses and normalized predictions to either option indices or option labels, depending on the benchmark annotation format. 

For PANDA, responses containing explicit null-valued JSON outputs
(e.g., \texttt{"isup\_grade": null}) were mapped to ISUP grade 0.
This handling was introduced because the model occasionally described slides as benign while emitting null rather than a numeric grade. This is due to the MultiPathQA PANDA prompt not including the description of 0 as benign ISUP.
Majority Vote predictions were evaluated using the aggregated majority label computed from per-patch predictions, while All-in-One predictions were evaluated directly from the single model response. Patch-level inference often yielded in refusal (the model stated that it cannot make a WSI-level prediction based on a single image). We excluded these instances from the majority vote aggregation.
Samples with failed API calls, empty outputs, or unparsable responses were retained in the evaluation and counted as incorrect predictions.

Performance was computed using the metric specified by each benchmark in MultiPathQA. Classification benchmarks used balanced accuracy, while VQA-style benchmarks used standard accuracy. Confidence intervals were estimated using nonparametric bootstrap resampling with 1000 bootstrap iterations and a fixed random seed. In addition to aggregate performance, the evaluation pipeline tracked per-sample correctness, parsing failures, API errors, token usage statistics, and reasoning token counts for each experimental configuration. Experimental variables, including magnification, patch size, patch count, and inference mode, were automatically extracted from result filenames to ensure consistency between the evaluated configurations and the original factorial design.

\subsection{External CPTAC validation set}
To assess whether the observed trends generalized beyond MultiPathQA, we constructed an additional external validation set from CPTAC whole-slide images available through the Imaging Data Commons. We queried CPTAC collections containing slide microscopy images and treated the CPTAC collection identifier as the primary diagnosis label. To reduce patient-level redundancy, we sampled at most one WSI per patient within each class. We then drew a stratified sample of 200 WSIs across CPTAC cancer types, using a fixed random seed for reproducibility. If a class contained fewer slides than requested, all available slides from that class were included, and the remaining sample budget was filled from the remaining eligible slides.

The CPTAC validation prompt followed the same format as the MultiPathQA TCGA primary diagnosis prompt, with the wording changed from ``histopathology image'' to ``pathology image'' because the CPTAC cohort includes acute myeloid leukemia cytology images. The answer options corresponded to CPTAC disease categories, including acute myeloid leukemia, breast invasive carcinoma, clear cell renal cell carcinoma, colorectal adenocarcinoma, glioblastoma, head and neck squamous cell carcinoma, lung squamous cell carcinoma, lung adenocarcinoma, ovarian serous cystadenocarcinoma, pancreatic adenocarcinoma, sarcoma, and uterine corpus endometrial carcinoma. Downloaded DICOM series were converted to TIFF format before processing with Trident, allowing the same patch extraction and inference pipeline to be used as in the main experiments. 15 slides could not be converted to TIFF or could not be read by Trident, resulting in a total of 185 valid images.

\section{Additional statistical analyses}
\textbf{Justification and implementation of two-way ANOVA.}
To quantify the relative contribution of each factor, we perform a two-way interaction ANOVA on the accuracy of each setting per benchmark.
We estimate effect sizes with omega squared ($\omega^2$):
\begin{equation}
\omega^2 = \max\!\left(0,\;
\frac{SS_{\text{effect}} - df_{\text{effect}} \cdot MS_{\text{error}}}
     {SS_{\text{total}} + MS_{\text{error}}} \right)
\end{equation}
We chose omega-squared to adjust for our small sample size.
We chose two-way rather than three-way interaction ANOVA to obtain more stable estimates of main effects. A three-way model would evaluate each effect conditional on all higher-order interactions, but would make isolating relevant effects more difficult due to our small sample size. We note that the reuse of slides, along with the derivation of Majority Vote subsets from shared runs, violates the independence assumption of ANOVA. Accordingly, we use it to identify which factors explain the most variance relative to one another, rather than to test for statistical significance. Additionally, the factorial design examines four factors and their pairwise interactions across five benchmarks, yielding many simultaneous comparisons. We do not apply formal multiple-comparison correction at this stage; instead, we treat Phase~1 as hypothesis-generating and require effects to replicate at full scale (Phase~2) and on held-out models and data (Phase~3) before drawing substantive conclusions. This three-phase design serves as our primary safeguard against overinterpreting noise in the underpowered subset analysis.

\subsection{Exploring field of view}
\label{app:fov}

Patch size affects both image resolution and field of view (FOV). At a fixed magnification, larger patch sizes capture a larger tissue region from the WSI while also increasing the final image resolution provided to the model. Consequently, patch size, magnification, and patch count jointly influence the total amount of tissue context visible during inference (see Figure~\ref{fig:same_coord_patches_fov} for visualization). 
To better understand these interactions, we first examined whether approximate FOV exhibited similar scaling behavior across inference modes.

\begin{figure}
    \centering
    \includegraphics[width=0.5\linewidth]{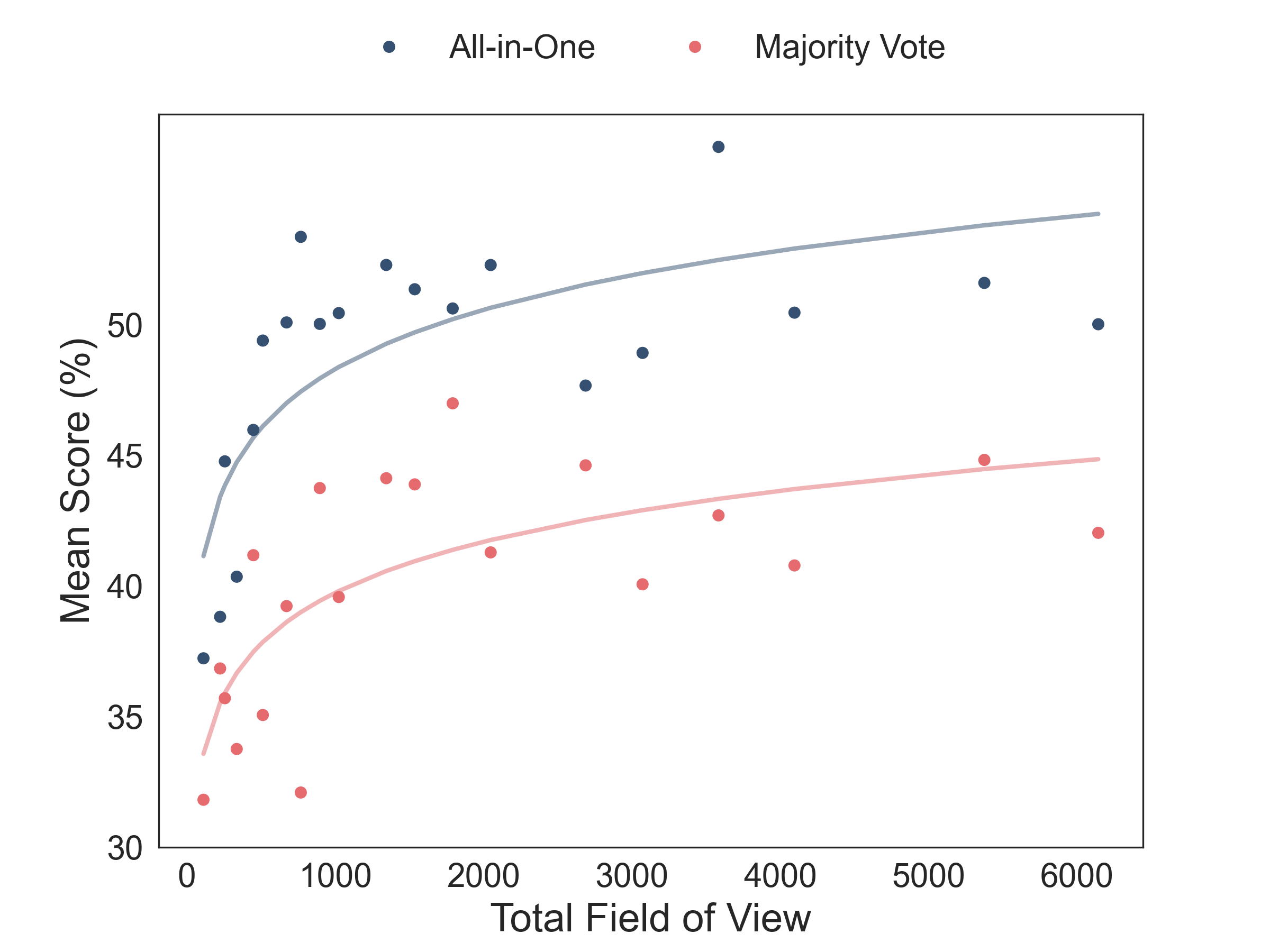}
    \caption{Total field of view,
        $(\text{Patch Size} / \text{Magnification}) \times \text{Patch Count}$.}
    \label{fig:FOV_subset}
\end{figure}

Figure~\ref{fig:FOV_subset} shows performance trends on the exploratory subset as a function of total FOV:
\[
\left(\frac{\text{Patch Size}}{\text{Magnification}}\right) \times \text{Patch Count}
\]
All-in-One and Majority Vote inference followed broadly similar trends with respect to FOV. However, All-in-One inference consistently achieved higher performance across most FOV ranges. Because of this consistent advantage, the subsequent analyses on the full MultiPathQA dataset focus exclusively on All-in-One inference.

\begin{figure}[h!]
    \centering
    \includegraphics[width=0.5\linewidth]{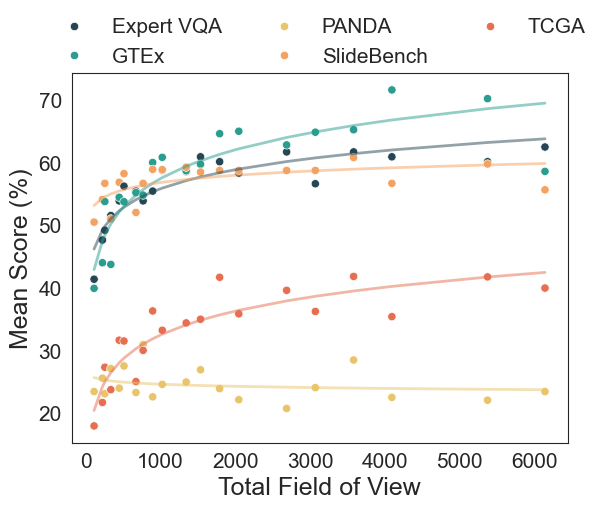}
    \caption{Scaling behavior as a function of total field of view across the full MultiPathQA dataset using All-in-One inference. Curves show benchmark-specific logarithmic fits.}
    \label{fig:fov_scatter_all}
\end{figure}

To investigate whether performance gains were primarily driven by increased tissue coverage, we next fit benchmark-specific logarithmic scaling curves relating total FOV to model performance across the full MultiPathQA dataset (Figure~\ref{fig:fov_scatter_all}). Across benchmarks, larger FOV generally correlated with improved performance, suggesting that broader tissue context is beneficial for many pathology tasks.

\begin{figure}[h!]
    \centering
    \includegraphics[width=\linewidth]{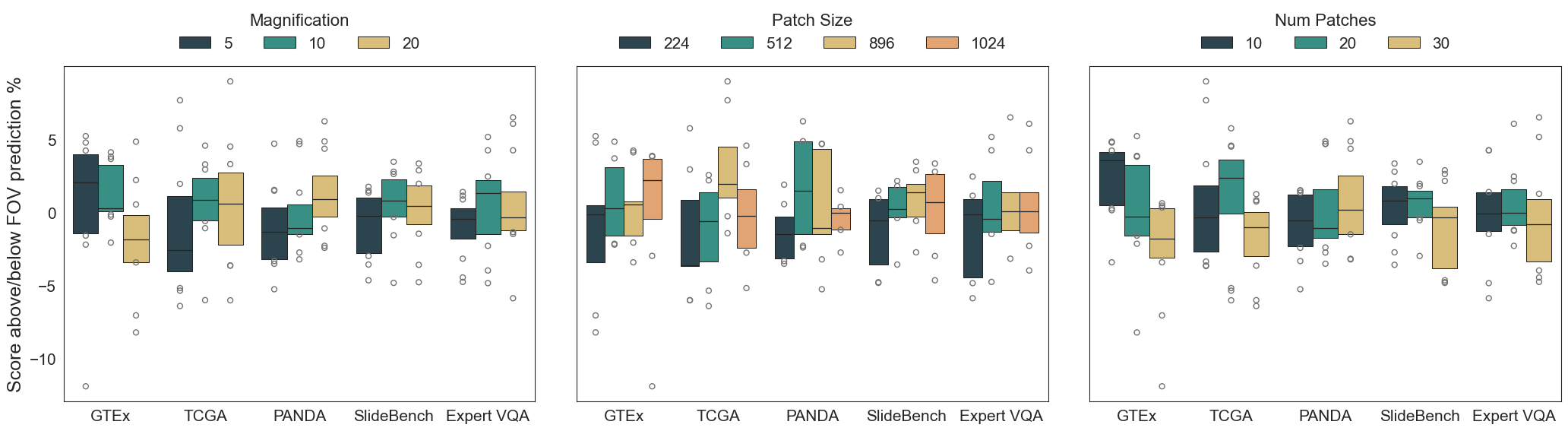}
    \caption{Performance relative to FOV-based prediction on the full MultiPathQA dataset. Boxen plots show the difference between observed performance and the value predicted from benchmark-specific logarithmic fits over total field of view. Results are grouped by magnification (left), patch size (middle), and number of patches (right).}
    \label{fig:FOV_predicted_diff}
\end{figure}

However, total field of view alone does not fully explain the observed scaling behavior. To disentangle the effects of tissue coverage from the independent effects of patch size, magnification, and patch count, we analyzed performance after subtracting the value predicted by the benchmark-specific FOV scaling fit (Figure~\ref{fig:FOV_predicted_diff}). Positive residuals indicate configurations that performed better than expected given their total field of view, whereas negative residuals indicate worse-than-expected performance.

The residual analyses reveal substantial benchmark-specific effects that persist even after controlling for FOV. GTEx and PANDA retain strong magnification-dependent behavior, with GTEx consistently favoring lower magnification (5$\times$) and PANDA favoring higher magnification (20$\times$). Similarly, TCGA and GTEx continue to exhibit sensitivity to patch size beyond what would be expected from tissue coverage alone. These trends closely mirror the ANOVA decomposition shown in Figure~\ref{fig:full_decomp}, where magnification and patch size explain substantial fractions of the total variance independently of FOV-related effects.

Together, these findings suggest that patch size, magnification, and patch count are not merely proxies for total tissue coverage. Instead, they independently alter the visual representation provided to the model. Importantly, configurations with large FOV are typically achieved through combinations of lower magnification, larger patch sizes, and higher patch counts, which themselves are often independently beneficial. As a result, FOV scaling trends partially conflate the effects of tissue coverage with favorable representation characteristics. Overall, the results indicate that while increased tissue coverage contributes to improved performance, benchmark-specific representation effects remain important even after controlling for FOV.

\section{Additional experimental results}

\subsection{Performance across configurations}
\label{app:all_configs}

\begin{figure}[h!]
    \centering
    \begin{subfigure}[t]{\linewidth}
        \centering
        \includegraphics[width=\linewidth]{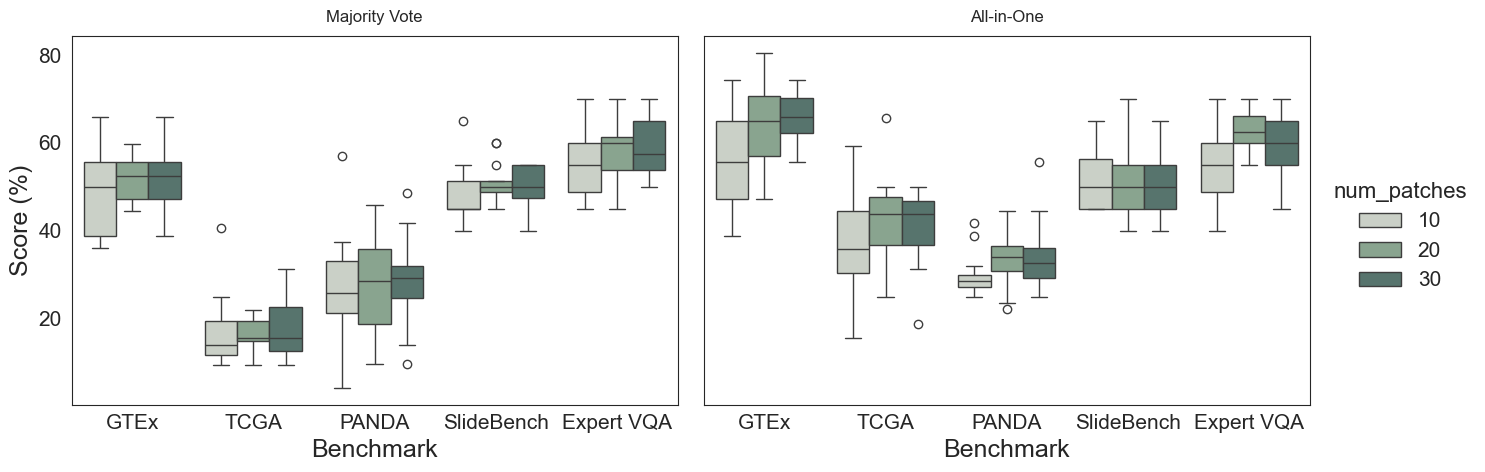}
        \caption{Patch Count}
        \label{fig:top5}
    \end{subfigure}
    \hfill
    \begin{subfigure}[t]{\linewidth}
        \centering
        \includegraphics[width=\linewidth]{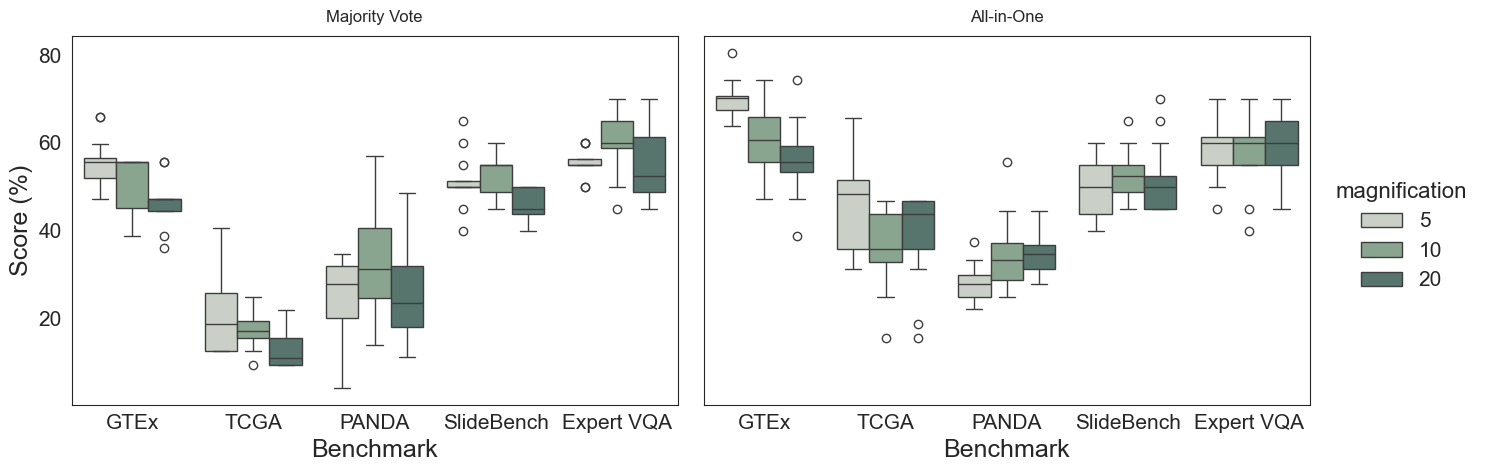}
        \caption{Magnification}
        \label{fig:bottom5_mag}
    \end{subfigure}
    \begin{subfigure}[t]{\linewidth}
        \centering
        \includegraphics[width=\linewidth]{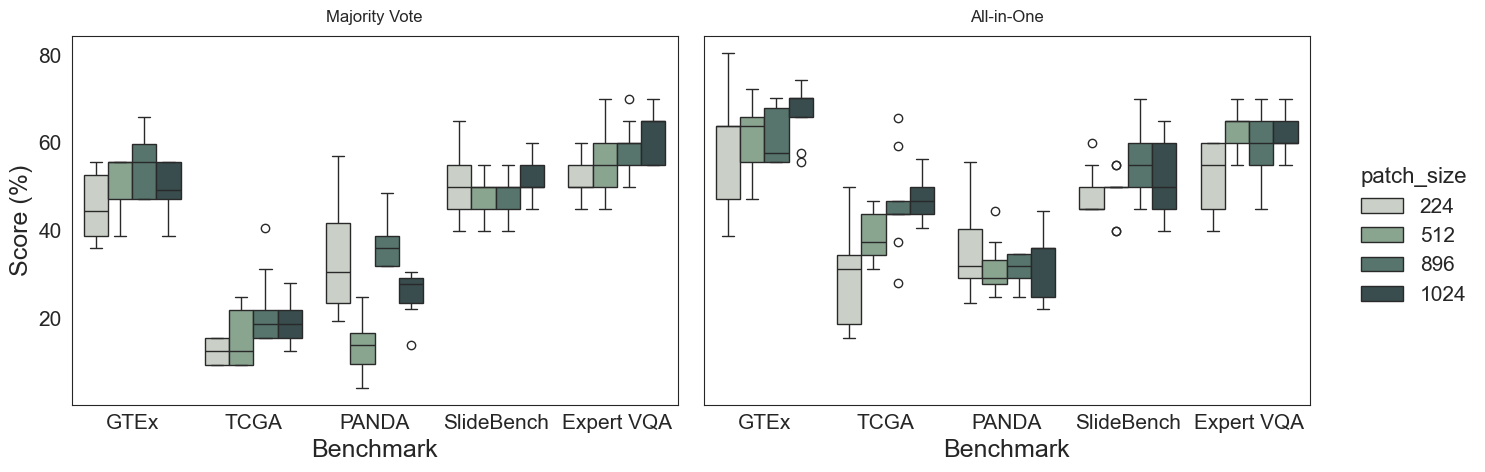}
        \caption{Patch size}
        \label{fig:bottom5}
    \end{subfigure}
    \caption{Performance across configurations.}
    \label{fig:supplement_fig1}
\end{figure}

Extending Figure \ref{fig:boxplots_subset_new}, we split the results for each factor by inference mode.
Figure~\ref{fig:supplement_fig1} summarizes performance across the main factorial axes: patch count, magnification, and patch size. Across most benchmarks, All-in-One inference outperforms Majority Vote, supporting the use of joint multi-patch context rather than independent per-patch predictions. The effect of patch count is benchmark dependent. GTEx generally benefits from additional patches, while the effect is weaker or less consistent for TCGA, PANDA, SlideBench, and ExpertVQA.

\subsection{Mixed-magnification ablation}
\label{app:mixed_mag}

Because benchmark performance varied across magnifications, we tested whether
combining multiple spatial scales in a single input could provide a more robust
representation. Using All-in-One inference, each whole-slide image was
represented by 30 total patches at 896\,px: 10 patches at 5$\times$, 10 at
10$\times$, and 10 at 20$\times$, all jointly provided to one GPT-5 instance.

As shown in Table~\ref{tab:mixed_mag_ablation}, mixed magnification did not consistently improve over fixed-magnification inputs. Instead, performance was typically intermediate relative to the 5$\times$, 10$\times$, and 20$\times$
settings. For TCGA and GTEx, the mixed configuration was below 5$\times$ but above 10$\times$ and 20$\times$. For PANDA, it improved over 5$\times$ and 10$\times$ but remained below 20$\times$. On SlideBench and ExpertVQA, the mixed setting underperformed all single-magnification baselines. Overall, we see that combining scales within a fixed patch budget tends to average the strengths of individual magnifications rather than produce additive gains.

\begin{table}[h!]
\centering
\caption{Mixed-magnification ablation on the full MultiPathQA benchmark using All-in-One inference. The mixed setting uses 30 total patches (896\,px), distributed evenly across magnifications: 10 patches each at 5$\times$, 10$\times$, and 20$\times$. Scores are balanced accuracy for classification tasks and accuracy for VQA tasks.}
\label{tab:mixed_mag_ablation}
\small
\setlength{\tabcolsep}{5pt}
\renewcommand{\arraystretch}{1.15}
\begin{tabular}{lccccc}
\toprule
Configuration & TCGA & GTEx & PANDA & SlideBench & ExpertVQA \\
\midrule
5$\times$ only  & $\mathbf{41.8 \pm 3.5}$ & $\mathbf{70.2 \pm 2.8}$ & $22.1 \pm 2.7$ & $58.2 \pm 3.5$ & $43.8 \pm 4.5$ \\
10$\times$ only & $39.6 \pm 3.3$ & $62.8 \pm 3.3$ & $20.8 \pm 2.7$ & $56.7 \pm 3.4$ & $43.8 \pm 4.5$ \\
20$\times$ only & $33.3 \pm 3.5$ & $56.8 \pm 2.5$ & $\mathbf{28.8 \pm 3.1}$ & $\mathbf{59.3 \pm 3.6}$ & $\mathbf{45.3 \pm 4.5}$ \\
Mixed (5$\times$+10$\times$+20$\times$)
               & $41.1 \pm 3.4$ & $66.4 \pm 3.1$ & $25.3 \pm 2.8$ & $53.3 \pm 3.6$ & $40.6 \pm 4.4$ \\
\bottomrule
\end{tabular}
\end{table}

\subsection{Improvement over prior patching Baseline}
\label{app:impr_over_prior}

To summarize the gains from our full ablation, we compared the prior baseline configuration (majority vote with 30 patches at 224\,px and 20$\times$ magnification) against the best single setting identified for each benchmark.
Across all five tasks, the best single-setting configuration substantially outperformed the prior baseline, with gains of +28.8 on TCGA, +33.5 on GTEx,
+9.5 on PANDA, +12.9 on SlideBench, and +24.2 on ExpertVQA
(Table~\ref{tab:improvement_over_prior_baseline}). The strongest configurations were task dependent: lower magnification and larger fields of view were most effective for TCGA and GTEx, whereas PANDA favored higher magnification, consistent with its reliance on finer cellular detail.

\begin{table}[h!]
\centering
\caption{Improvement over the prior baseline on MultiPathQA. The prior baseline
uses majority vote with 30 patches at 224\,px and 20$\times$ magnification. The
comparison uses the best single setting identified independently for each
benchmark. Scores are balanced accuracy for classification tasks and accuracy
for VQA tasks. Two settings on ExpertVQA reached the same accuracy so we report the one with lower standard deviation.}
\label{tab:improvement_over_prior_baseline}
\small
\setlength{\tabcolsep}{5pt}
\renewcommand{\arraystretch}{1.15}
\begin{tabular}{l l c c c}
\toprule
Benchmark & Best setting & Prior baseline & Best score & $\Delta$ \\
\midrule
TCGA &
10p, 896\,px, 5$\times$ &
$15.1 \pm 2.0$ &
$43.9 \pm 3.9$ &
$+28.8$ \\

GTEx &
20p, 1024\,px, 5$\times$ &
$38.1 \pm 2.7$ &
$71.6 \pm 2.9$ &
$+33.5$ \\

PANDA &
30p, 512\,px, 20$\times$ &
$21.5 \pm 2.5$ &
$31.0 \pm 2.9$ &
$+9.5$ \\

SlideBench &
20p, 896\,px, 10$\times$ &
$49.0 \pm 3.6$ &
$61.9 \pm 3.4$ &
$+12.9$ \\

ExpertVQA &
30p, 512\,px, 10$\times$ &
$39.1 \pm 4.3$ &
$63.3 \pm 4.3$ &
$+24.2$ \\
\bottomrule
\end{tabular}
\end{table}
\section{Cost, token usage, and API reliability}
\label{app:cost_token_error}

\subsection{Token usage and cost}
\label{app:cost_token}

GPT-5 pricing during experimentation was 1.25 USD per million input tokens and 10 USD per million output tokens (including reasoning tokens). Azure Batch API requests are priced at approximately half the cost of standard API requests. Consequently, Majority Vote inference in the exploratory subset experiments benefited from lower effective pricing despite substantially higher token consumption. Similarly, the full factorial experiment on the complete MultiPathQA dataset was executed entirely using the Batch API, reducing overall inference cost.

Token usage and estimated inference costs for the exploratory subset experiments are summarized in Table~\ref{tab:token_usage_subset}, while aggregate statistics for the full factorial experiment on the complete MultiPathQA dataset are reported in Table~\ref{tab:token_usage_full}. Reported averages are computed per evaluated WSI--configuration result, where a result corresponds to a single inference call for a specific whole-slide image under a particular magnification, patch size, patch count, and inference mode configuration.

For Majority Vote inference, token accounting includes only the maximum patch-count configuration per experimental setting. Patch-level predictions were generated once using the largest requested patch count (30 patches), and results for smaller patch-count settings (10 and 20 patches) were reconstructed using nested subsets of these predictions without additional API calls. Consequently, the number of evaluated results differs between inference modes in Table~\ref{tab:token_usage_subset}, and Majority Vote token totals do not include duplicated charges for subsetted configurations.

Additional token usage and cost statistics for model experiments used in Phase 3 are summarized in Table~\ref{tab:additional_model_token_usage}. These experiments include the prior GPT-5 Majority Vote baseline protocol (224\,px, 20$\times$ magnification, 30 patches) evaluated on the full MultiPathQA benchmark suite, as well as All-in-One inference experiments using Gemini 3 Flash and Qwen 3.5 Plus through OpenRouter. Estimated costs for Gemini 3 Flash during time of inference were 0.376 USD per million input tokens and 3 USD per million output tokens. Qwen 3.5 Plus costs were 0.433 USD per million input tokens and 2.40 USD per million output tokens.

\begin{table*}[h!]
\centering
\small
\caption{
Token usage and estimated inference cost statistics for the exploratory subset experiments. Average values are computed per evaluated WSI--configuration result. Estimated costs reflect the pricing schemes used during experimentation. All-in-One inference used standard GPT-5 API pricing, whereas Majority Vote inference used Batch API pricing. For Majority Vote, token usage is counted only for the maximum patch-count run, since lower patch-count settings were reconstructed from nested subsets of the same per-patch predictions.
}
\resizebox{\linewidth}{!}{
\begin{tabular}{llrrrrrrrrr}
\toprule
Benchmark & Method &
N &
Total Tokens &
Input Tokens &
Output Tokens &
Avg Tokens / Result &
Avg Input / Result &
Avg Output / Result &
Cost (USD) &
Cost / Result \\
\midrule

GTEx & All-in-One &
720 &
6,772,417 &
6,130,350 &
642,067 &
9,406 &
8,514 &
892 &
13.86 &
0.0193 \\

GTEx & Majority Vote &
240 &
7,627,474 &
4,088,684 &
3,538,790 &
31,781 &
17,036 &
14,745 &
20.25 &
0.0844 \\

TCGA & All-in-One &
720 &
7,199,059 &
6,278,850 &
920,209 &
9,999 &
8,721 &
1,278 &
17.05 &
0.0237 \\

TCGA & Majority Vote &
240 &
11,613,141 &
5,401,894 &
6,211,247 &
48,388 &
22,508 &
25,880 &
34.43 &
0.1435 \\

PANDA & All-in-One &
720 &
6,606,131 &
5,182,380 &
1,423,751 &
9,175 &
7,198 &
1,977 &
20.72 &
0.0288 \\

PANDA & Majority Vote &
240 &
13,542,637 &
3,368,988 &
10,173,649 &
56,428 &
14,037 &
42,390 &
52.97 &
0.2207 \\

SlideBench & All-in-One &
720 &
6,810,042 &
5,962,479 &
847,563 &
9,458 &
8,281 &
1,177 &
15.93 &
0.0221 \\

SlideBench & Majority Vote &
240 &
8,226,101 &
3,799,557 &
4,426,544 &
34,275 &
15,831 &
18,444 &
24.51 &
0.1021 \\

ExpertVQA & All-in-One &
720 &
6,825,574 &
6,111,396 &
714,178 &
9,480 &
8,488 &
992 &
14.78 &
0.0205 \\

ExpertVQA & Majority Vote &
240 &
7,034,220 &
3,793,368 &
3,240,852 &
29,309 &
15,806 &
13,504 &
18.58 &
0.0774 \\

\midrule

Total &
Combined &
4,800 &
82,256,796 &
50,117,946 &
32,138,850 &
17,137 &
10,441 &
6,696 &
231.83 &
0.0483 \\

\bottomrule
\end{tabular}
}
\label{tab:token_usage_subset}
\end{table*}

\begin{table*}[h!]
\centering
\small
\caption{
Token usage and estimated inference cost statistics for the full factorial experiments using All-in-One inference.
Average values are computed per evaluated WSI--configuration result.
Estimated costs reflect the pricing scheme used during experimentation.
}
\resizebox{\linewidth}{!}{
\begin{tabular}{llrrrrrrrrr}
\toprule
Benchmark & Method &
N &
Total Tokens &
Input Tokens &
Output Tokens &
Avg Tokens / Result &
Avg Input / Result &
Avg Output / Result &
Cost (USD) &
Cost / Result \\
\midrule

GTEx & All-in-One &
6,876 &
68,392,552 &
57,830,348 &
10,562,204 &
9,947 &
8,410 &
1,536 &
88.95 &
0.0129 \\

TCGA & All-in-One &
7,956 &
86,789,290 &
68,348,770 &
18,440,520 &
10,909 &
8,591 &
2,318 &
134.92 &
0.0170 \\

PANDA & All-in-One &
7,092 &
72,237,325 &
51,155,160 &
21,082,165 &
10,186 &
7,213 &
2,973 &
137.38 &
0.0194 \\

SlideBench & All-in-One &
6,984 &
72,816,352 &
58,321,982 &
14,494,370 &
10,426 &
8,351 &
2,075 &
108.92 &
0.0156 \\

ExpertVQA & All-in-One &
4,608 &
48,018,741 &
38,803,584 &
9,215,157 &
10,421 &
8,421 &
2,000 &
70.33 &
0.0153 \\

\midrule

Total &
Combined &
33,516 &
348,254,260 &
274,459,844 &
73,794,416 &
10,391 &
8,189 &
2,202 &
540.51 &
0.0161 \\

\bottomrule
\end{tabular}
}
\label{tab:token_usage_full}
\end{table*}

\begin{table*}[h!]
\centering
\small
\caption{
Token usage and estimated inference cost statistics for additional model experiments. GPT-5 experiments used the prior Majority Vote baseline protocol (224\,px, 20$\times$ magnification, 30 patches). Gemini 3 Flash and Qwen 3.5 Plus experiments were performed through OpenRouter using All-in-One inference. Average values are computed per evaluated WSI--configuration result. Estimated costs were calculated using provider pricing rates at the time of experimentation.
}
\resizebox{\linewidth}{!}{
\begin{tabular}{lrrrrrrrrr}
\toprule
Benchmark &
N &
Total Tokens &
Input Tokens &
Output Tokens &
Avg Tokens &
Avg Input &
Avg Output &
Cost (USD) &
Cost / Result \\
\midrule

\multicolumn{10}{c}{\textbf{GPT-5 Prior Protocol (Azure Batch API)}} \\
\midrule

GTEx &
191 &
8,167,987 &
2,072,088 &
6,095,899 &
42,764 &
10,849 &
31,916 &
31.77 &
0.1664 \\

PANDA &
197 &
17,228,613 &
2,229,822 &
14,998,791 &
87,455 &
11,319 &
76,136 &
76.39 &
0.3878 \\

TCGA &
221 &
14,214,013 &
3,586,414 &
10,627,599 &
64,317 &
16,228 &
48,089 &
55.38 &
0.2506 \\

ExpertVQA &
128 &
4,896,969 &
1,216,908 &
3,680,061 &
38,258 &
9,507 &
28,750 &
19.16 &
0.1497 \\

SlideBench &
194 &
9,357,260 &
1,934,516 &
7,422,744 &
48,233 &
9,972 &
38,262 &
38.32 &
0.1975 \\

\midrule

Total &
931 &
53,864,842 &
11,039,748 &
42,825,094 &
57,857 &
11,858 &
45,999 &
221.03 &
0.2374 \\

\midrule
\midrule

\multicolumn{10}{c}{\textbf{Gemini 3 Flash (OpenRouter)}} \\
\midrule

GTEx &
382 &
10,895,702 &
10,393,849 &
501,853 &
28,523 &
27,209 &
1,314 &
5.54 &
0.0145 \\

PANDA &
394 &
11,609,141 &
10,562,035 &
1,047,106 &
29,465 &
26,807 &
2,658 &
7.24 &
0.0184 \\

TCGA &
442 &
12,794,190 &
12,110,842 &
683,348 &
28,946 &
27,400 &
1,546 &
6.75 &
0.0153 \\

ExpertVQA &
256 &
7,387,152 &
6,953,916 &
433,236 &
28,856 &
27,164 &
1,692 &
4.00 &
0.0156 \\

SlideBench &
394 &
11,268,901 &
10,474,690 &
794,211 &
28,601 &
26,586 &
2,016 &
6.45 &
0.0164 \\

\midrule

Total &
1,868 &
53,955,086 &
50,495,332 &
3,459,754 &
28,884 &
27,032 &
1,852 &
29.97 &
0.0160 \\

\midrule
\midrule

\multicolumn{10}{c}{\textbf{Qwen 3.5 Plus (OpenRouter)}} \\
\midrule

GTEx &
382 &
4,532,433 &
3,403,242 &
1,129,191 &
11,865 &
8,909 &
2,956 &
4.07 &
0.0107 \\

PANDA &
394 &
6,304,122 &
3,325,278 &
2,978,844 &
16,000 &
8,440 &
7,561 &
8.48 &
0.0215 \\

TCGA &
442 &
5,866,733 &
3,940,152 &
1,926,581 &
13,273 &
8,914 &
4,359 &
6.20 &
0.0140 \\

ExpertVQA &
256 &
3,376,226 &
2,208,143 &
1,168,083 &
13,188 &
8,626 &
4,563 &
3.69 &
0.0144 \\

SlideBench &
394 &
5,255,102 &
3,281,104 &
1,973,998 &
13,338 &
8,328 &
5,010 &
6.05 &
0.0154 \\

\midrule

Total &
1,868 &
25,334,616 &
16,157,919 &
9,176,697 &
13,562 &
8,650 &
4,913 &
28.49 &
0.0153 \\

\bottomrule
\end{tabular}
}
\label{tab:additional_model_token_usage}
\end{table*}

\subsection{Refusal rates}
\label{app:refusal}
A small number of requests were rejected by Azure content filtering systems due to medical image content policy violations. These cases were logged and excluded from downstream analysis rather than retried.
API refusal statistics are summarized in Tables~\ref{tab:subset_refusals} and~\ref{tab:full_dataset_refusals}. Across both the exploratory subset and full factorial experiments, refusal rates were low overall ($<0.1\%$ total). Most refusals occurred in the SlideBench benchmark, likely reflecting the broader diversity of medical image content and visual presentation styles contained within the dataset.

\begin{table}[h!]
\centering
\small
\caption{
Azure/OpenAI API refusal statistics for the exploratory subset factorial experiments.
Refusals correspond to requests rejected by content filtering systems.
}
\begin{tabular}{lrrrr}
\toprule
Benchmark & Total & Successful & Refusals & Refusal Rate (\%) \\
\midrule

GTEx & 960 & 960 & 0 & 0.00 \\

PANDA & 960 & 960 & 0 & 0.00 \\

TCGA & 960 & 960 & 0 & 0.00 \\

ExpertVQA & 960 & 960 & 0 & 0.00 \\

SlideBench & 960 & 957 & 3 & 0.31 \\

\midrule

Total & 4,800 & 4,797 & 3 & 0.06 \\

\bottomrule
\end{tabular}
\label{tab:subset_refusals}
\end{table}

\begin{table}[h!]
\centering
\small
\caption{
Azure/OpenAI API refusal statistics for the full factorial experiment on the complete MultiPathQA dataset.
Refusals correspond to requests rejected by content filtering systems.
}
\begin{tabular}{lrrrr}
\toprule
Benchmark & Total & Successful & Refusals & Refusal Rate (\%) \\
\midrule

GTEx & 7,067 & 7,065 & 2 & 0.03 \\

PANDA & 7,289 & 7,286 & 3 & 0.04 \\

TCGA & 8,177 & 8,173 & 4 & 0.05 \\

ExpertVQA & 4,736 & 4,732 & 4 & 0.08 \\

SlideBench & 7,178 & 7,160 & 18 & 0.25 \\

\midrule

Total & 34,447 & 34,416 & 31 & 0.09 \\

\bottomrule
\end{tabular}
\label{tab:full_dataset_refusals}
\end{table}

\section{Patch visualizations}
\label{app:patch_visualizations}

\begin{figure}[h!]
    \centering
    \includegraphics[width=0.9\linewidth]{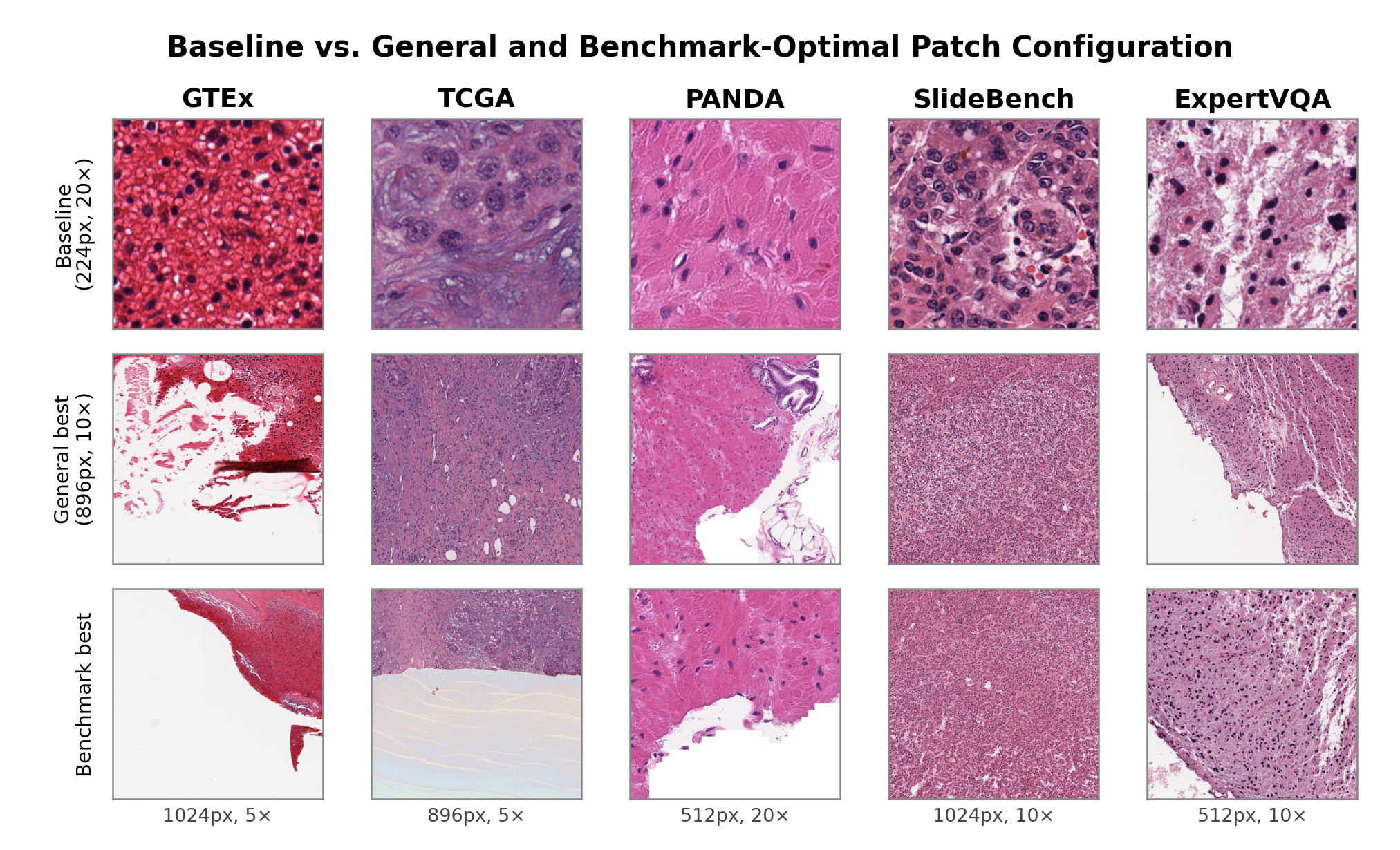}
    \caption{Representative example patches comparing the prior baseline configuration (224\,px at 20$\times$ magnification; top row), the benchmark-agnostic optimized configuration (896\,px at 10$\times$ magnification; middle row), and the best-performing configuration per benchmark (bottom row). Patches were sampled independently from valid patch coordinate sets for each configuration, using the same WSI within each benchmark.}
    \label{fig:baseline_vs_optimal_patch}
\end{figure}

Figure~\ref{fig:baseline_vs_optimal_patch} shows representative example patches from three configurations: the prior baseline setting (224\,px at 20$\times$ magnification), the benchmark-agnostic optimized setting (896\,px at 10$\times$ magnification), and the best-performing configuration identified separately for each benchmark. For each benchmark, patches were sampled from the same WSI but independently from the valid patch coordinate set for each configuration.

These examples qualitatively illustrate how patching configurations change both field of view and visual representation. The baseline setting emphasizes highly localized cellular morphology, whereas the optimized configurations expose substantially larger tissue context. For GTEx and TCGA, the benchmark-specific optima use lower magnification and larger patch sizes, producing patches that capture broader tissue architecture and spatial organization. SlideBench similarly benefits from larger contextual fields, although at an intermediate magnification of 10$\times$.

PANDA shows a different pattern: its benchmark-specific optimum retains high magnification while increasing patch size relative to the baseline. This preserves fine glandular and nuclear detail while modestly increasing contextual coverage. ExpertVQA also favors an intermediate magnification setting, suggesting that both local morphology and larger tissue structure contribute to performance.

The benchmark-agnostic optimized setting visually resembles the benchmark-specific optima for several tasks, particularly those favoring larger tissue context. For PANDA and ExpertVQA, however, the benchmark-agnostic setting provides less cellular detail than the task-specific optimum, consistent with the stronger role of fine morphology in these benchmarks.

\begin{figure}[h!]
    \centering
    \includegraphics[width=0.9\linewidth]{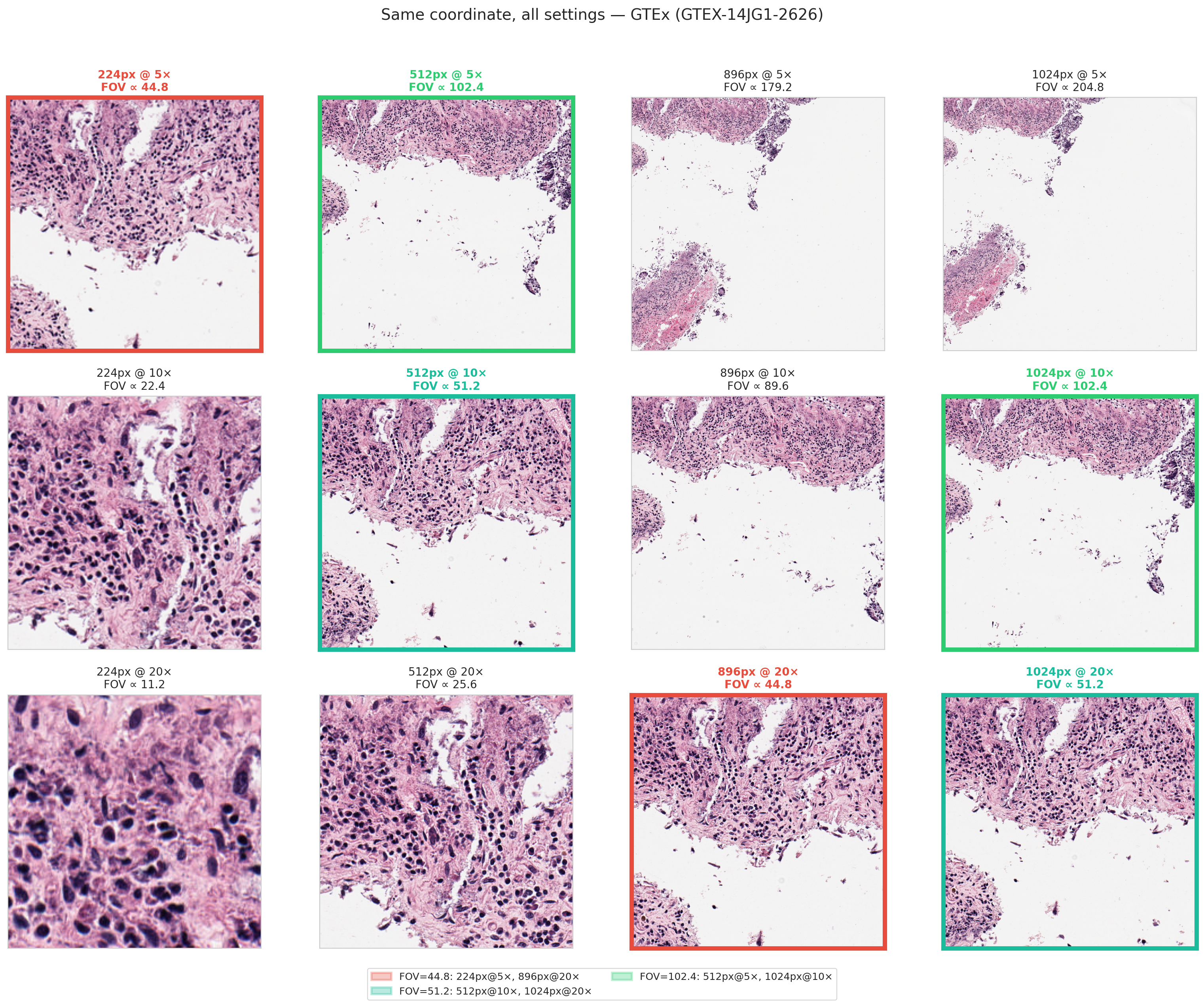}
    \caption{Representative patches extracted from the same spatial coordinate under different magnification and patch-size configurations. Colored outlines indicate configurations with approximately matched field of view (FOV). While several combinations produce similar tissue coverage, the resulting visual representations have different image sizes/resolution.}
    \label{fig:same_coord_patches_fov}
\end{figure}

To supplement the Field-of-View analysis, Figure~\ref{fig:same_coord_patches_fov} provides a qualitative illustration of how equivalent approximate field-of-view values can arise from different combinations of magnification and patch size. Although these configurations may cover similar tissue areas, they produce substantially different visual representations and effective image resolutions.

%%%%%%%%%%%%%%%%%%%%%%%%%%%%%%%%%%%%%%%%%%%%%%%%%%%%%%%%%%%%
\end{document}